%% file: main.tex
\documentclass[10pt,twocolumn,letterpaper]{article}

\usepackage[pagenumbers]{iccv} 


\definecolor{iccvblue}{rgb}{0.21,0.49,0.74}
\usepackage[pagebackref,breaklinks,colorlinks,allcolors=iccvblue]{hyperref}

\usepackage{amsmath}
\usepackage{multirow}
\usepackage{booktabs}  
\usepackage{makecell}  
\usepackage[capitalize]{cleveref}
\usepackage{url}

\def\paperID{12327} 
\def\confName{ICCV}
\def\confYear{2025}

\title{Event-boosted Deformable 3D Gaussians for Dynamic Scene Reconstruction}

\author{Wenhao Xu \hspace{8pt} Wenming Weng \hspace{8pt} Yueyi Zhang \hspace{8pt} Ruikang Xu \hspace{8pt} Zhiwei Xiong \\
University of Science and Technology of China \\
{\tt\small \{wh-xu, wmweng, xurk\}@mail.ustc.edu.cn, \{zhyuey, zwxiong\}@ustc.edu.cn}}

\begin{document}

\twocolumn[{
\renewcommand\twocolumn[1][]{#1}
\maketitle

\begin{center}
    \includegraphics[width=1.\linewidth]{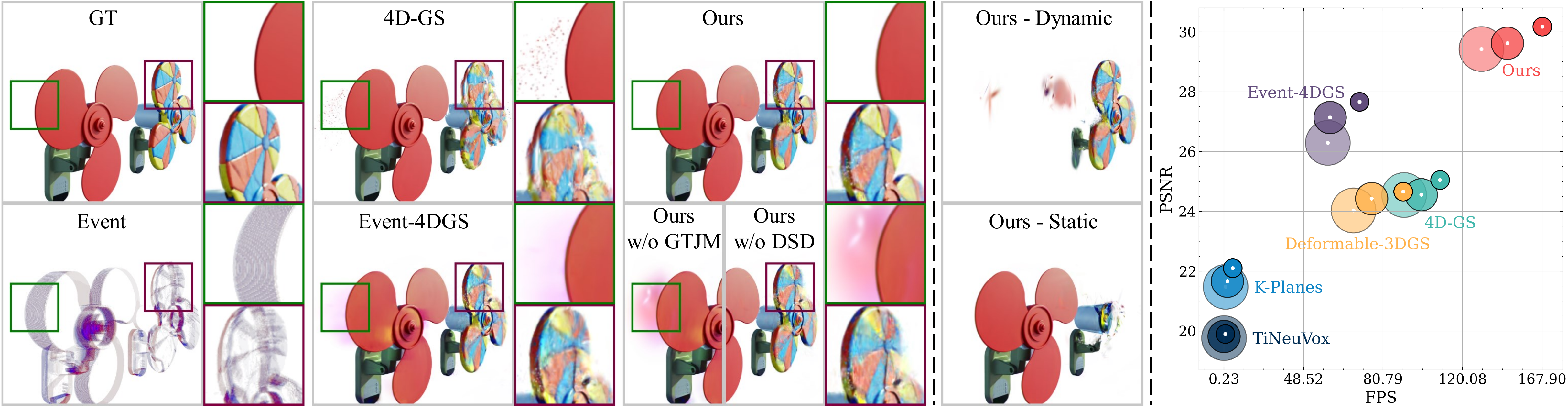}
    \vspace{-2mm}
    \captionsetup{type=figure}
    \caption{
    \textbf{Left}: 
    Quality comparison with the baselines 4D-GS \cite{wu20244d}, Event-4DGS (the event-extended version of \cite{yang2024deformable}), and our variants, highlighting the superior rendering quality of our method. Our GS-threshold joint modeling (GTJM) effectively eliminates event-induced artifacts (particularly the purple haze), while our dynamic-static decomposition (DSD) improves the quality of dynamic regions.
    \textbf{Middle}:
    Separate rendering of dynamic and static Gaussians from our DSD.
    \textbf{Right}: The scatter plot illustrates our method’s ability to achieve both high fidelity and fast rendering, where dot radii correspond to different resolutions (400$\times$400, 600$\times$600, and 800$\times$800).
    }
    \label{fig:teaser}
\end{center}
}]

\begin{abstract}

Deformable 3D Gaussian Splatting (3D-GS) is limited by missing intermediate motion information due to the low temporal resolution of RGB cameras. To address this, we introduce the first approach combining event cameras, which capture high-temporal-resolution, continuous motion data, with deformable 3D-GS for dynamic scene reconstruction. We observe that threshold modeling for events plays a crucial role in achieving high-quality reconstruction. Therefore, we propose a GS-Threshold Joint Modeling strategy, creating a mutually reinforcing process that greatly improves both 3D reconstruction and threshold modeling. Moreover, we introduce a Dynamic-Static Decomposition strategy that first identifies dynamic areas by exploiting the inability of static Gaussians to represent motions, then applies a buffer-based soft decomposition to separate dynamic and static areas. This strategy accelerates rendering by avoiding unnecessary deformation in static areas, and focuses on dynamic areas to enhance fidelity. Additionally, we contribute the first event-inclusive 4D benchmark with synthetic and real-world dynamic scenes, on which our method achieves state-of-the-art performance.
\end{abstract}

\section{Introduction}
\label{sec:intro}


Dynamic scene reconstruction and novel view synthesis are essential for immersive applications in virtual/augmented reality and entertainment \cite{wu20244d, yang2024deformable, luiten2023dynamic, liang2023gaufre, shaw2024swings, yang2023real, zhang2024gaussiancube}. While Neural Radiance Fields (NeRF) \cite{mildenhall2021nerf, barron2021mip, verbin2022ref} offer unprecedented photorealism, they are constrained by slow training and rendering speeds. Despite recent advances in optimization techniques \cite{cao2023hexplane, fang2022fast, fridovich2022plenoxels, fridovich2023k}, real-time rendering remains elusive. Recently, 3D Gaussian Splatting (3D-GS) \cite{kerbl20233d} addresses this limitation through efficient differentiable rasterization, yet existing dynamic extensions \cite{luiten2024dynamic, yang2024deformable, wu20244d, yang2023real} are constrained by inherent limitations of RGB cameras, including low frame rates and motion blur.

In this paper, we present event-boosted deformable 3D Gaussians for dynamic scene reconstruction. Event cameras \cite{gallego2020event, li2015design}, with their microsecond-level temporal resolution, can provide continuous motion information and near-infinite viewpoints that traditional RGB cameras often fail to capture. These advantages make event cameras particularly valuable for dynamic scene reconstruction.

However, integrating events into 3D scene reconstruction faces new challenges. Specifically, event supervision for 3D-GS relies on an accurate event generation model \cite{klenk2023nerf, cannici2024mitigating}, where the threshold undergoes complex variations across polarity, space, and time \cite{delbruck2020v2e,klenk2023nerf}. Previous methods \cite{rudnev2023eventnerf, klenk2023nerf, qi2023e2nerf, cannici2024mitigating, ma2023deformable, deguchi2024e2gs, yu2024evagaussians, xiong2024event3dgs, wu2024ev} adopt a constant threshold, yet this simplification significantly degrades the quality of event supervision. While recent works \cite{hwang2023ev, low2023robust} attempt to model threshold variations using event data alone, they achieve limited success due to the inherent binary nature of events, which only indicate brightness change directions. To address this challenge, we propose a novel GS-threshold joint modeling strategy. 
First, we leverage the brightness change values from RGB frames to supervise threshold optimization.
Second, since the sparsity of RGB frames weakens supervision, we use 3D-GS rendered results as pseudo-intermediate frames to enhance the supervision.
This finally creates a mutually reinforcing process where RGB-optimized threshold enables better event supervision for 3D-GS, while improved 3D-GS in turn provides accurate geometric constraints for threshold refinement.

Furthermore, we observe that existing dynamic 3D-GS methods inefficiently use dynamic Gaussians solely to model both static and dynamic regions \cite{wu20244d, yang2024deformable, luiten2023dynamic, lu20243d, huang2024sc, guo2024motion, bae2024per}. This unified treatment leads to reduced rendering speed, wasted deformation field capacity, and degraded reconstruction quality. While some methods have explored dynamic-static decomposition, they are limited by either inaccurate dynamic Gaussians initialization \cite{liang2023gaufre} or constraints in multi-view scenarios \cite{shaw2024swings}. To address these limitations, we propose a novel dynamic-static decomposition strategy that first identifies dynamic regions based on the inherent inability of static Gaussians to represent motions, and then employs a buffer-based soft decomposition to adaptively search for the optimal decomposition boundary. This decomposition not only accelerates rendering by eliminating unnecessary deformation computations in static regions, but also enhances reconstruction quality by focusing the deformation field exclusively on dynamic regions.

Our main contributions can be summarized as follows:
\begin{itemize}
\item We present the first method integrating event cameras with deformable 3D-GS for dynamic scene reconstruction, enabling high-fidelity and fast rendering.
\item We propose a novel GS-threshold joint modeling strategy that combines RGB-assisted initial estimation with GS-boosted refinement, creating a mutually reinforcing process that significantly improves both threshold modeling and 3D reconstruction.
\item We introduce an effective dynamic-static decomposition strategy that not only accelerates rendering through selective deformation computation but also enhances reconstruction quality by focusing on dynamic regions.
\item We contribute the first event-inclusive 4D benchmark with synthetic and real-world dynamic scenes, on which our method achieves state-of-the-art performance.
\end{itemize}

\section{Related Work}
\label{sec:related}

\noindent\textbf{Neural Rendering for Dynamic Scenes.}
Neural rendering techniques have revolutionized dynamic scene reconstruction in recent years. Pioneering works like D-NeRF \cite{pumarola2021d} and Nerfies \cite{park2021nerfies} extend Neural Radiance Fields \cite{mildenhall2021nerf} through deformation field, mapping observations into a canonical space for modeling non-rigid motion. Despite their impressive reconstruction quality, these methods are constrained by extensive computational demands due to dense MLP evaluations during training and rendering. Various acceleration strategies have been proposed to address these limitations. K-Planes \cite{fridovich2023k} introduces an efficient explicit representation using six feature planes, while Tensor4D \cite{shao2023tensor4d} and DTensoRF \cite{jang2022d} employ tensor decomposition techniques to achieve compact spatiotemporal encoding. A significant breakthrough came with 3D Gaussian Splatting (3D-GS) \cite{kerbl20233d}, which leverages efficient differentiable rasterization for real-time rendering. This advancement has spawned several dynamic scene extensions, including 4D-GS \cite{wu20244d}, Deformable-3DGS \cite{yang2024deformable}, and related works \cite{luiten2023dynamic, shaw2024swings},  which successfully achieve real-time rendering for high-quality dynamic scene reconstruction.

\noindent\textbf{Event-based Neural Rendering.}
The integration of neural representations with event-based 3D reconstruction has emerged as a promising research direction. Pioneering works like EventNeRF \cite{rudnev2023eventnerf}, E-NeRF \cite{klenk2023nerf}, and Ev-NeRF \cite{hwang2023ev} first demonstrated the potential of pure event-based static scene reconstruction, albeit with different assumptions about event camera characteristics. To further improve reconstruction quality, E2NeRF \cite{qi2023e2nerf} and Ev-DeblurNeRF \cite{cannici2024mitigating} incorporated blurry RGB images alongside events. A significant milestone was achieved by DE-NeRF \cite{ma2023deformable}, which pioneered the combination of events and RGB frames for dynamic scene reconstruction. The recent advent of 3D Gaussian Splatting \cite{kerbl20233d} has catalyzed new developments in event-based methods. While Ev-GS \cite{wu2024ev} adapted the pure event-based paradigm to 3D-GS, subsequent works including E2GS \cite{deguchi2024e2gs}, EaDeblur-GS \cite{weng2024eadeblur}, and Event3DGS \cite{xiong2024event3dgs} primarily addressed deblurring challenges. To the best of our knowledge, our work is the first to integrate events with deformable 3D-GS for dynamic scene reconstruction. \textit{Our novel method and benchmark, specifically tailored for events and dynamic scenes,} highlight significant and independent contributions that set our work apart.

\section{Method}
\label{sec:method}

\begin{figure}[t!]
  \centering
  \includegraphics[width=\linewidth]{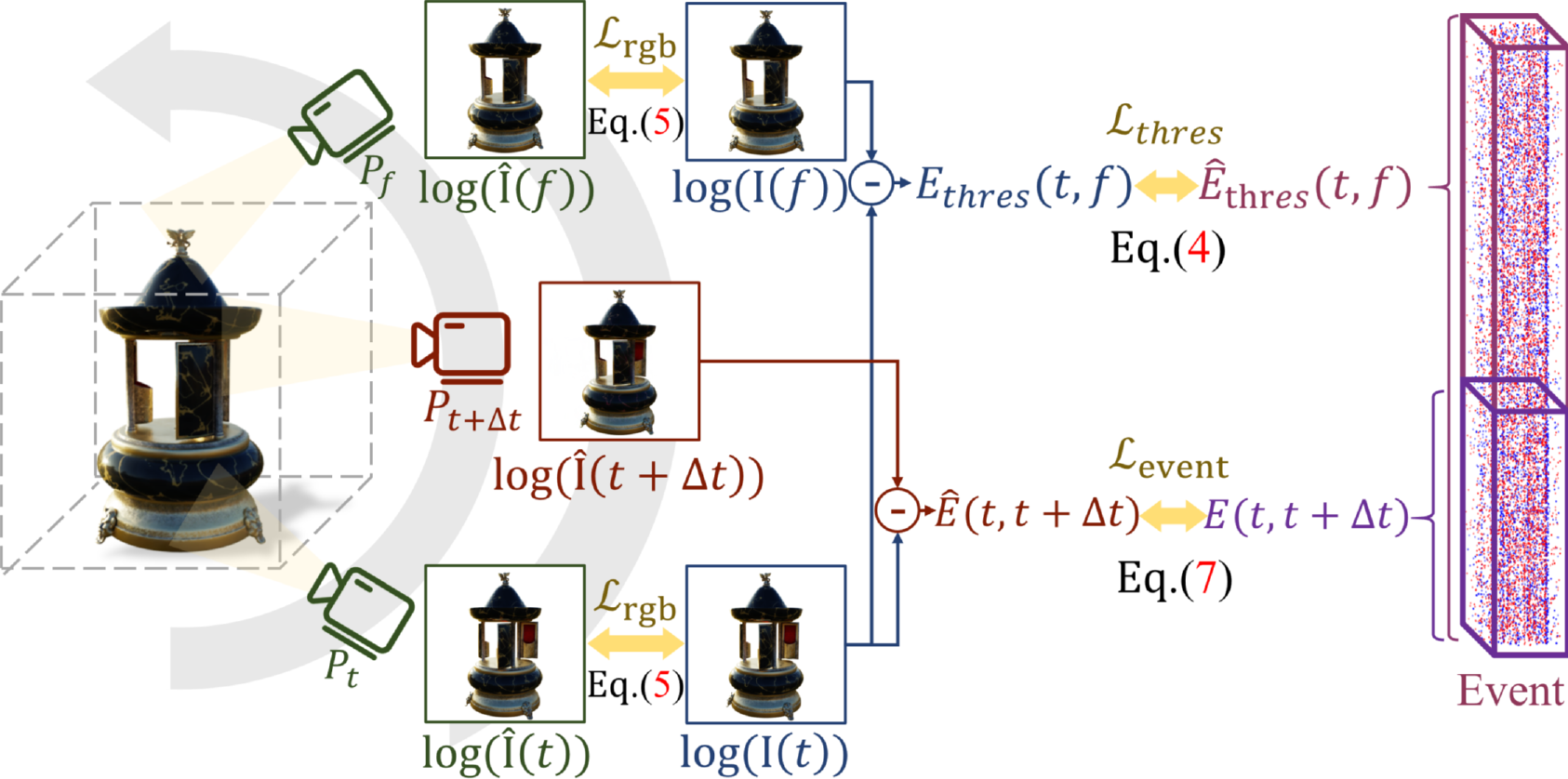}
  \caption{Overview of GS-threshold joint modeling strategy. $\mathcal{L}_{rgb}$ optimizes 3D-GS, $\mathcal{L}_{thres}$ optimizes the threshold, and $\mathcal{L}_{event}$ jointly optimizes both 3D-GS and threshold.}
  \label{fig:pipeline}
\end{figure}

\subsection{Event Cameras for 3D-GS}
\noindent\textbf{3D Gaussian Splatting Preliminary.}
3D Gaussian Splatting (3D-GS) \cite{kerbl20233d} represents a scene as anisotropic 3D Gaussians, each characterized by a covariance matrix $\Sigma$ and center position $\mathbf{\mu}$: $GS(\mathbf{x})=e^{-\frac{1}{2}(\mathbf{x}-\mathbf{\mu})^{T}\sum ^{-1}(\mathbf{x}-\mathbf{\mu})}$.
The covariance matrix $\Sigma$ is parameterized using scaling matrix $S$ and rotation matrix $R$ to ensure positive semi-definiteness: $\Sigma=RSS^{T}R^{T}$.
Each Gaussian is further defined by spherical harmonic coefficients $\mathcal{C}$ and opacity $\sigma$. Final pixel colors $c$ are computed using differentiable tile-based rasterization
\begin{equation}
c=\sum_{i\in N}c_{i}\alpha_{i}\prod_{j=1}^{i-1}(1-\alpha_{j}),
\end{equation}
where $c_{i}$ denotes spherical harmonic color, and $\alpha_{i}$ combines opacity $\sigma$ with projected $GS(\mathbf{x})$.

\noindent\textbf{Event/RGB Rendering Loss.} 
Event cameras \cite{gallego2020event, li2015design} are novel sensors, that asynchronously capture pixel-wise brightness changes with microsecond-level temporal resolution. Their high temporal precision enables capturing crucial motions between RGB frames and provides near-infinite viewpoint supervision, making them ideal for monocular dynamic scene reconstruction.

Each event is represented as $e_{x,y}(\tau)=p\delta(\tau)$, where $(x,y)$ is pixel position, $\tau$ is timestamp, $p\in \left\{+1,-1 \right\}$ indicates brightness change direction relative to threshold $C$, and $\delta (t)$ is a unit integral impulse function. 
Omitting pixel subscripts, the brightness change over interval $\bigtriangleup t$ can be formulated as
\begin{equation}
E(t,t+\bigtriangleup t)=\int_{t}^{t+\bigtriangleup t}C\cdot e(\tau)d\tau.
\label{eq:event_integration}
\end{equation}
This change can also be estimated from rendered brightness:
\begin{equation}
\hat{E}(t,t+\bigtriangleup t):=\log(\hat{I}(t+\bigtriangleup t))-\log(I(t)).
\end{equation}
where $\hat{I}$ and $I$ respectively denote 3D-GS rendered and ground truth brightnesses. The event rendering loss is
\begin{equation}
\mathcal{L}_{event}=\left \| E(t,t+\bigtriangleup t) - \hat{E} (t,t+\bigtriangleup t)\right \|_{2}^{2}.
\label{eq:event_loss}
\end{equation}
Similarly, we utilize the RGB rendering loss \cite{kerbl20233d} combining L1 and D-SSIM losses as
\begin{equation}
\mathcal{L}_{rgb}=(1-\lambda_{s})\left \| \hat{I}(t)-I(t)\right \|_{1}+\lambda_{s}\mathcal{L}_{D-SSIM}(\hat{I}(t),I(t)),
\label{eq:rgb_loss}
\end{equation}
where $\lambda_{s}$ is a weighting factor that controls the balance. 3D-GS is optimized to minimize $\mathcal{L}_{event}$ and $\mathcal{L}_{rgb}$ jointly. 
\begin{equation}
GS^{\ast}=\arg \underset{GS}{\min}(\mathcal{L}_{event}+\mathcal{L}_{rgb}).
\end{equation}

\begin{figure}[t!]
  \centering
  \includegraphics[width=1.0\linewidth]{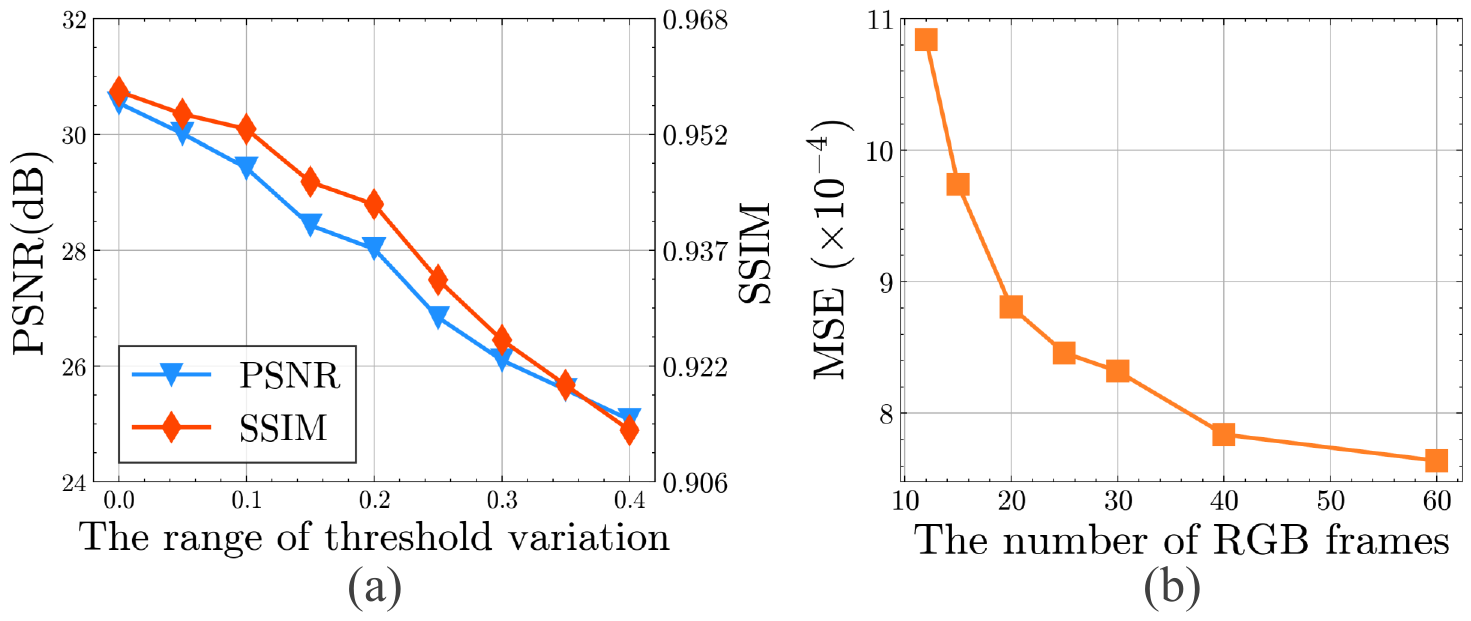}
  \vspace{-6mm}
  \caption{(a) The effect of different ranges of threshold variation for 3D reconstruction.
  (b) The effect of different number of RGB frames for threshold estimation.
  }
  \label{fig:vary}
  \vspace{-2mm}
\end{figure}

\subsection{GS-threshold Joint Modeling }
As shown in \cref{eq:event_integration}, the threshold $C$ critically affects event integration and supervision quality. While existing methods \cite{rudnev2023eventnerf, klenk2023nerf, qi2023e2nerf, cannici2024mitigating, ma2023deformable} typically assume a constant threshold, real event cameras exhibit threshold variations across polarity, space, and time \cite{delbruck2020v2e,klenk2023nerf}. \cref{fig:vary} (a) demonstrates how increasing threshold variation significantly degrades reconstruction quality. To model threshold variations, we propose a \textit{GS-threshold joint modeling (GTJM) strategy} (see \cref{fig:pipeline}), consisting of RGB-assisted threshold estimation and GS-boosted threshold refinement.

\noindent\textbf{RGB-assisted Threshold Estimation.}
Recent works \cite{hwang2023ev, low2023robust} attempt to model threshold variations solely from event data but face significant limitations due to the difficulty of inferring brightness changes from events, which only indicate the direction of change.
We propose to leverage brightness change values from RGB frames for robust threshold estimation.
Given two RGB frames $I(t)$ and $I(f)$ at times $t$ and $f$, we define the threshold modeling loss as
\begin{equation}
\mathcal{L}_{thres}=\left \| E_{thres}(t,f) - \hat{E}_{thres} (t,f)\right \|_{2}^{2},
\label{eq:thres_loss}
\end{equation}
where $\hat{E}_{thres}(t,f):=\int_{t}^{f}\hat{C}\cdot e(\tau)d\tau$, $E_{thres}(t,f)=log(I(f))-log(I(t))$.
In practice, we adopt a simple yet fast way to compute $\hat{E}_{thres}(t,f)$.
We first accumulate events to obtain event count maps \cite{gehrig2019end} $ECM_{t,f} \in\mathbb{R}^{B\times P \times H \times W}$, where $B$ denotes time bins and $P=2$ corresponds to the event polarity. Using learnable threshold parameters $\hat{C}_{t,f} \in\mathbb{R}^{B\times P \times H \times W}$, we compute
\begin{equation}
\hat{E}_{thres}(t,f)=\sum_{b=1}^{B} \sum_{p=1}^{P} \left ( ECM_{t,f} \odot \hat{C}_{t,f} \right )_{b,p,:,:},
\end{equation}
where threshold $\hat{C}_{t,f}$ is optimized by minimizing \cref{eq:thres_loss} in an end-to-end manner.

We observe that \textit{accurate threshold modeling improves 3D reconstruction quality}. As shown in ``TM for 3D Rec.'' in \cref{tab:Thres_modeling_table}, our RGB-assisted threshold optimization approach benefits threshold estimation, thus significantly enhancing 3D-GS reconstruction quality and achieving a 2.17 dB PSNR improvement.

\begin{table}[t]
  \caption{Step-by-step validation of mutual boosting between threshold modeling (TM) and 3D reconstruction (3D Rec.). Abbreviations: ``Fro.'': ``Frozen''; ``Ft.'': ``Fine-tuning''. Note that TM is evaluated by MSE between the estimated and GT thresholds from the simulator.}
  \vspace{-2mm}
  \centering
    \setlength{\tabcolsep}{5pt}
  \renewcommand{\arraystretch}{1}
  \resizebox{1.0\linewidth}{!}{
  \begin{tabular}{cc|c|cc|c}
    \toprule
    \multicolumn{3}{c|}{TM for 3D Rec.} & \multicolumn{3}{c}{3D Rec. for TM} \\
    \cmidrule(lr){1-3} \cmidrule(lr){3-6}
    
    Stage1 & Stage2 & 3D Rec. & Stage1 & Stage2 & TM \\
    \cmidrule(lr){1-2} \cmidrule(lr){3-3} \cmidrule(lr){4-5} \cmidrule(lr){6-6}
    TM & \makecell[c]{3D Rec.\\(Fro. $\hat{C}$)} & PSNR$\uparrow$ & 3D Rec. & \makecell[c]{TM\\(Fro. GS)} & \makecell[c]{MSE$\downarrow$\\($\times10^{-4}$)} \\
    \cmidrule(lr){1-2} \cmidrule(lr){3-3} \cmidrule(lr){4-5} \cmidrule(lr){6-6}
    $ \times $ & $ \checkmark $ & 24.46 & $ \times $ & $ \checkmark $ & 8.317 \\
    $ \checkmark $ & $ \checkmark $ & 26.63 & $ \checkmark $ & $ \checkmark $ & 7.077 \\
    \midrule
    \midrule

    \multicolumn{6}{c}{Joint TM and 3D Rec. Optimization} \\
    \cmidrule(lr){1-6} 
    
    Stage1 & Stage2 & \multicolumn{3}{c|}{3D Rec.} & TM \\
    \cmidrule(lr){1-2} \cmidrule(lr){3-5} \cmidrule(lr){6-6}
    TM & \makecell[c]{3D Rec. \& TM\\ (Ft. $\hat{C}$ \& Ft. GS)} & \multicolumn{3}{c|}{PSNR$\uparrow$} & \makecell[c]{MSE$\downarrow$\\($\times10^{-4}$)} \\
    \cmidrule(lr){1-2} \cmidrule(lr){3-5} \cmidrule(lr){6-6}
    $ \checkmark $ & $ \checkmark $ & \multicolumn{3}{c|}{\textbf{28.01}} & \textbf{6.322} \\

    \bottomrule
  \end{tabular}
  }
  \label{tab:Thres_modeling_table}
  \vspace{-4mm}
\end{table}

\noindent\textbf{GS-boosted Threshold Refinement.} 
While RGB frames facilitate threshold estimation, their effectiveness is constrained by low frame rate. As illustrated in \cref{fig:vary} (b), sparse RGB frames lead to longer integration intervals, reducing supervision quality and threshold estimation accuracy. To overcome this limitation, we found that once a 3D-GS is trained first by \cref{eq:event_loss} and \cref{eq:rgb_loss}, it can be used to render intermediate frames as additional pseudo-supervision.
Specifically, we freeze the trained 3D-GS and reuse \cref{eq:event_loss} to enhance \cref{eq:thres_loss} for optimizing threshold $\hat{C}$
\begin{equation}
\hat{C}^{\ast}=\arg \underset{\hat{C}}{\min}(\mathcal{L}_{thres}+\mathcal{L}_{event}).
\end{equation}


We observe that \textit{the incorporation of 3D-GS significantly enhances threshold estimation accuracy}. 
The underlying reason is that events may provide unreliable supervision in some regions with inaccurate thresholds or noise, whereas 3D-GS can correct these errors via geometric consistency. 
As demonstrated in ``3D Rec. for TM'' part in \cref{tab:Thres_modeling_table}, using trained and frozen 3D-GS for threshold modeling substantially reduces MSE, leading to more precise threshold estimation.

\noindent\textbf{Joint Threshold and GS Optimization.} 
Having demonstrated the mutual benefits between threshold modeling and 3D reconstruction, we propose jointly optimizing both threshold $\hat{C}$ and 3D Gaussians $GS$ through
\begin{equation}
\hat{C}^{\ast},GS^{\ast}=\arg \underset{\hat{C}, GS}{\min}(\mathcal{L}_{thres}+\mathcal{L}_{event}+\mathcal{L}_{rgb}).
\end{equation}

We observe that \textit{this joint optimization enables a beneficial cycle where optimized thresholds enhance event supervision for 3D-GS, while improved 3D-GS refines threshold estimates through geometric consistency}. 
As shown in ``Joint TM and 3D Rec. Optimization'' part in \cref{tab:Thres_modeling_table}, this approach achieves superior threshold modeling and reconstruction quality.

In summary, our optimization proceeds in two stages: first optimizing the threshold using $\mathcal{L}_{thres}$, then jointly optimizing both threshold and 3D-GS using all three losses $\mathcal{L}_{thres}$, $\mathcal{L}_{event}$, and $\mathcal{L}_{rgb}$.

\begin{figure}[t!]
  \centering
  \includegraphics[width=0.9\linewidth]{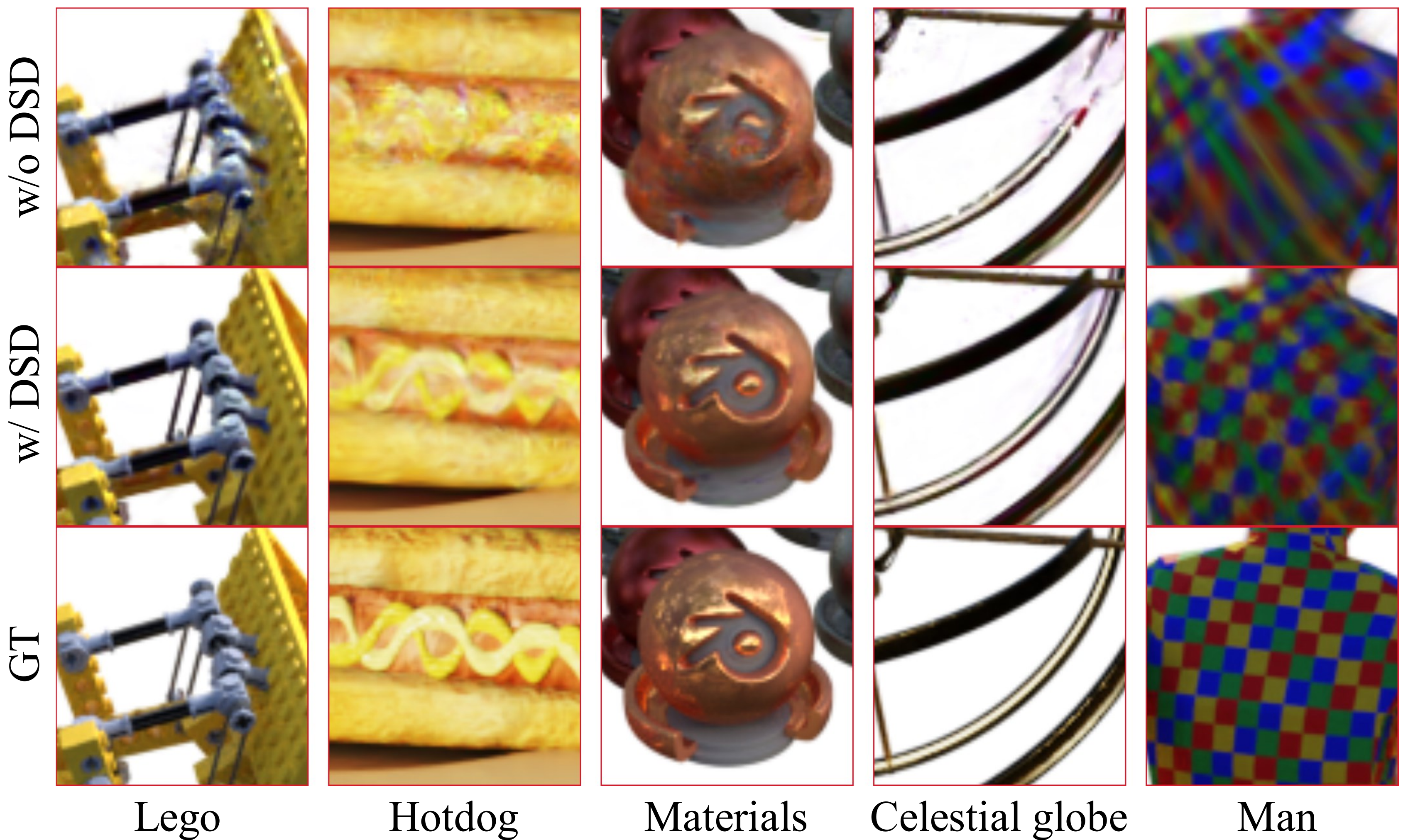}
  \caption{The effect of dynamic-static decomposition strategy, which improves the rendering quality of dynamic regions.}
  \label{fig:wo_decomp_comparison}
  \vspace{-4mm}
\end{figure}

\subsection{Dynamic-static Decomposition}
\label{sec:dSd}

Dynamic scenes typically contain substantial static regions (e.g., tables, walls) that require no deformation. Unlike existing methods \cite{wu20244d, yang2024deformable, luiten2023dynamic, lu20243d, huang2024sc, guo2024motion, bae2024per} that use dynamic Gaussians throughout, we separately model dynamic and static regions with corresponding Gaussian types. This decomposition offers dual benefits: accelerated rendering by bypassing deformation field computation for static Gaussians, and enhanced deformation fidelity through focused MLP capacity optimization for dynamic regions, as demonstrated in \cref{fig:wo_decomp_comparison}.

\begin{figure*}[t!]
  \centering
  \includegraphics[width=1.\textwidth]{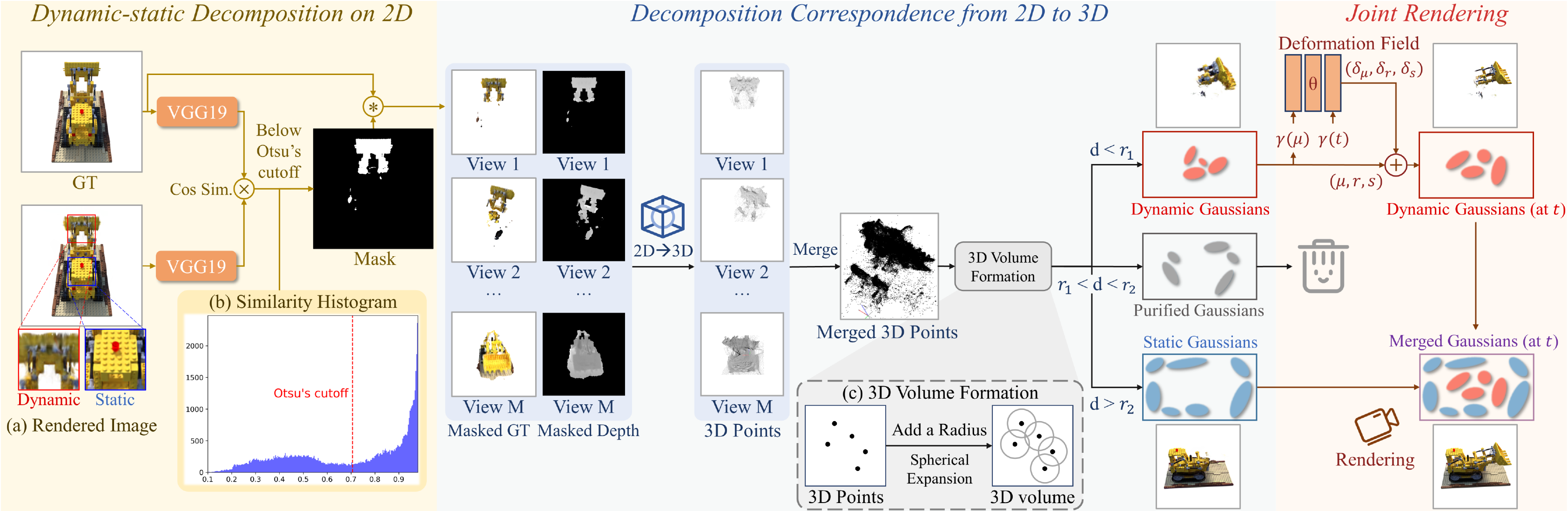}
  \caption{Overview of dynamic-static decomposition strategy. First, we decompose dynamic and static regions in 2D images based on the inherent inability of static Gaussians to represent motions. Next, we establish a correspondence to extend 2D decomposition to 3D Gaussians. Finally, the decomposed dynamic and static Gaussians are jointly rendered to reconstruct the complete dynamic scene.}
  \label{fig:decomp_pipeline}
  \vspace{-4mm}
\end{figure*}

The key challenge lies in accurately initializing dynamic Gaussians in dynamic regions and static Gaussians in static regions. We address this through a proposed \textit{dynamic-static decomposition (DSD) strategy}, as illustrated in \cref{fig:decomp_pipeline}.

\noindent\textbf{Dynamic-static Decomposition on 2D.} 
We leverage the inherent inability of static Gaussians in representing motion to decompose dynamic and static regions in 2D images. During the first 3k iterations, we perform scene reconstruction using only static Gaussians for initialization. This naturally results in poor reconstruction in dynamic regions while achieving high fidelity in static areas (illustrated in \cref{fig:decomp_pipeline} (a)). This distinct performance difference enables decomposition of training images into dynamic and static regions.

Specifically, using a pretrained VGG19 \cite{simonyan2014very} network $\mathcal{F}_{\phi}$, we extract multi-scale features from both rendered image $\hat{I}(t)$ and ground truth $I(t)$. The cosine similarities computed at each scale are upsampled to a uniform resolution and averaged to generate a fused similarity map
\begin{equation}
Sim=\sum_{l} Up \left ( \frac{\mathcal{F}_{\phi }^{l}(\hat{I}(t))\cdot\mathcal{F}_{\phi }^{l}(I(t)) }{\left \| \mathcal{F}_{\phi }^{l}(\hat{I}(t)) \right \|  \left \| \mathcal{F}_{\phi }^{l}(I(t)) \right \| }\right ),
\end{equation}
where $\mathcal{F}_{\phi }^{l}(\cdot )$ represents the $l$-th layer output of VGG19, and $Up(\cdot)$ indicates bilinear upsampling. The histogram of the resulting similarity map exhibits a bimodal distribution as shown in \cref{fig:decomp_pipeline} (b), enabling dynamic region mask generation through Otsu's \cite{otsu1975threshold} method
\begin{equation}
Mask=\mathbf{1}_{Sim<Otsu(Sim)},
\end{equation}
where $\mathbf{1}_{\left \{ \cdot  \right \} }$ denotes the indicator function, which returns 1 if the condition is true. The mask is then multiplied by the ground truth image to extract the dynamic region.

\noindent\textbf{Decomposition Correspondence from 2D to 3D.}
To extend 2D dynamic-static decomposition to 3D Gaussians, we establish view-independent correspondences by leveraging depth information from 3D-GS rendering. By unprojecting pixels from masked dynamic regions across multiple views and merging the resulting 3D points, we obtain a comprehensive representation of dynamic regions in 3D space.


Next, we map merged points to dynamic Gaussians based on spatial proximity. Each point expands spherically with radius $r$ to form a 3D volume (\cref{fig:decomp_pipeline} (c)), initially classifying enclosed Gaussians as dynamic and others as static. To overcome potential decomposition inaccuracies and radius sensitivity, we implement a \textit{buffer-based soft decomposition strategy} using two radii, $r_1$ and $r_2$. Gaussians within $r_1$ are marked as dynamic, beyond $r_2$ as static, while those between are pruned to create a buffer zone. This strategy enables 3D-GS to optimize decomposition boundaries through adaptive density control (ADC) \cite{kerbl20233d}, enhancing both rendering quality and speed. As demonstrated in \cref{fig:buffer_size}, the strategy also exhibits improved robustness to radius parameter selection.

It should be noted that, our DSD method is performed only once during the entire training process and requires only about one minute, introducing minimal overhead to the training pipeline.

\noindent\textbf{Joint Rendering of Dynamic and Static Gaussians.} With the decomposed dynamic and static Gaussians, we jointly render the entire dynamic scene. Particularly, a deformation field \cite{yang2024deformable} learns to map dynamic Gaussians from canonical space to arbitrary time. Taking time t and the center position $\boldsymbol{\mu}$ of dynamic Gaussians as inputs, the deformation field outputs the displacement of their position $\boldsymbol{\delta_{\mu}}$, rotation $\boldsymbol{\delta_{r}}$, and scaling $\boldsymbol{\delta_{s}}$
\begin{equation}
(\boldsymbol{\delta_{\mu}}, \boldsymbol{\delta_{r}},\boldsymbol{ \delta_{s}})=\mathcal{F}_{\theta}(\gamma(sg(\boldsymbol{\mu})),\gamma(t)),
\end{equation}
Where $sg(\cdot )$ indicates a stop-gradient operation and $\gamma (\cdot )$ denotes the positional encoding \cite{yang2024deformable}. Then, the deformed dynamic gaussians can be addressed as
\begin{equation}
(\boldsymbol{\mu} ',\boldsymbol{r}',\boldsymbol{s}')=(\boldsymbol{\mu} +\boldsymbol{\delta_{\mu}},\boldsymbol{r}+\boldsymbol{\delta_{r}},\boldsymbol{s}+ \boldsymbol{\delta_{s}}).
\end{equation}
Finally, static Gaussians bypass the deformation field and merge with the deformed dynamic Gaussians as inputs to the rasterizer, enabling high-frame-rate dynamic rendering.

\section{Experiment}
\label{sec:experiment}

\subsection{Experimental Settings}


\begin{table*}[t]
  \caption{Quantitative results on our synthetic dataset. Event-4DGS is an extension of Deformable-3DGS \cite{yang2024deformable} by incorporating events.}
  \vspace{-3mm}
  \centering
    \setlength{\tabcolsep}{4pt}
  \renewcommand{\arraystretch}{1}
  \resizebox{1.0\linewidth}{!}{
  \begin{tabular}{c|cccc|cccc|cccc|cccc}
    \toprule

    \multirow{2}{*}{\centering Method} & \multicolumn{4}{c|}{Lego} & \multicolumn{4}{c|}{Hotdog} & \multicolumn{4}{c|}{Materials} & \multicolumn{4}{c}{Music box} \\
    
    \cmidrule(lr){2-5} \cmidrule(lr){6-9} \cmidrule(lr){10-13} \cmidrule(lr){14-17}
    
    & PSNR$\uparrow$ & SSIM$\uparrow$ & LPIPS$\downarrow$ & FPS$\uparrow$ & PSNR$\uparrow$ & SSIM$\uparrow$ & LPIPS$\downarrow$ & FPS$\uparrow$ & PSNR$\uparrow$ & SSIM$\uparrow$ & LPIPS$\downarrow$ & FPS$\uparrow$ & PSNR$\uparrow$ & SSIM$\uparrow$ & LPIPS$\downarrow$ & FPS$\uparrow$ \\

    \cmidrule(lr){1-1} \cmidrule(lr){2-5} \cmidrule(lr){6-9} \cmidrule(lr){10-13} \cmidrule(lr){14-17}

    3D-GS \cite{kerbl20233d} & 23.60 & 0.918 & 0.088 & \textbf{223} & 30.01 & 0.951 & 0.064 & \textbf{260} & 28.07 & 0.967 & 0.061 & \textbf{262} & 19.20 & 0.905 & 0.122 & \textbf{239} \\
    TiNeuVox \cite{fang2022fast} & 22.39 & 0.891 & 0.071 & 0.53 & 30.81 & 0.953 & 0.035 & 0.49 & 26.63 & 0.938 & 0.054 & 0.52 & 20.45 & 0.831 & 0.152 & 0.62 \\
    K-Planes \cite{fridovich2023k} & 24.55 & 0.931 & \underline{0.035} & 2.34 & 31.36 & 0.958 & \underline{0.016} & 2.35 & 30.62 & 0.976 & 0.009 & 2.29 & 20.77 & 0.858 & 0.071 & 2.39 \\
    4D-GS \cite{wu20244d} & 26.30 & 0.937 & 0.072 & 104 & 33.48 & 0.965 & 0.052 & 132 & 30.40 & 0.979 & 0.054 & 111 & 24.06 & 0.937 & 0.071 & 64 \\
    Deformable-3DGS \cite{yang2024deformable} & 23.79 & 0.923 & 0.053 & 73 & 32.91 & 0.962 & 0.017 & 132 & 34.00 & 0.986 & \underline{0.004} & 91 & 22.08 & 0.924 & 0.052 & 51 \\
    Event-4DGS & \underline{28.00} & \underline{0.943} & 0.040 & 54 & \underline{34.61} & \underline{0.969} & 0.019 & 96 & \underline{35.60} & \underline{0.989} & 0.006 & 74 & \underline{28.58} & \underline{0.950} & \underline{0.043} & 42 \\
    Ours & \textbf{31.85} & \textbf{0.967} & \textbf{0.018} & \underline{189} & \textbf{36.15} & \textbf{0.974} & \textbf{0.013} & \underline{241} & \textbf{38.02} & \textbf{0.993} & \textbf{0.003} & \underline{240} & \textbf{30.78} & \textbf{0.963} & \textbf{0.029} & \underline{92} \\

    \midrule
    
    \multirow{2}{*}{\centering Method} & \multicolumn{4}{c|}{Celestial globe} & \multicolumn{4}{c|}{Fan} & \multicolumn{4}{c|}{Water wheel} & \multicolumn{4}{c}{Man} \\
    
    \cmidrule(lr){2-5} \cmidrule(lr){6-9} \cmidrule(lr){10-13} \cmidrule(lr){14-17}
    
    & PSNR$\uparrow$ & SSIM$\uparrow$ & LPIPS$\downarrow$ & FPS$\uparrow$ & PSNR$\uparrow$ & SSIM$\uparrow$ & LPIPS$\downarrow$ & FPS$\uparrow$ & PSNR$\uparrow$ & SSIM$\uparrow$ & LPIPS$\downarrow$ & FPS$\uparrow$ & PSNR$\uparrow$ & SSIM$\uparrow$ & LPIPS$\downarrow$ & FPS$\uparrow$ \\

    \cmidrule(lr){1-1} \cmidrule(lr){2-5} \cmidrule(lr){6-9} \cmidrule(lr){10-13} \cmidrule(lr){14-17}

    3D-GS \cite{kerbl20233d} & 19.05 & 0.915 & 0.110 & \textbf{182} & 21.26 & 0.891 & 0.118 & \textbf{270} & 19.43 & 0.887 & 0.109 & \textbf{215} & 20.79 & 0.870 & 0.114 & \textbf{210} \\
    TiNeuVox \cite{fang2022fast} & 13.62 & 0.736 & 0.290 & 0.62 & 19.90 & 0.889 & 0.107 & 0.53 & 17.03 & 0.850 & 0.147 & 0.56 & 22.81 & 0.887 & 0.071 & 0.50 \\
    K-Planes \cite{fridovich2023k} & 15.49 & 0.857 & 0.088 & 2.46 & 22.10 & 0.909 & 0.062 & 2.38 & 20.96 & 0.920 & \underline{0.046} & 2.41 & 21.02 & 0.857 & 0.073 & 2.33 \\
    4D-GS \cite{wu20244d} & 20.97 & 0.942 & 0.072 & 52 & 25.05 & 0.936 & 0.080 & 108 & 20.96 & 0.917 & 0.077 & 77 & 23.80 & 0.914 & 0.076 & 64 \\
    Deformable-3DGS \cite{yang2024deformable} & 23.07 & \underline{0.962} & \underline{0.036} & 41 & 24.66 & 0.929 & 0.051 & 90 & 20.79 & 0.912 & 0.051 & 43 & 23.06 & 0.906 & \underline{0.051} & 37 \\
    Event-4DGS & \underline{24.30} & 0.948 & 0.045 & 36 & \underline{27.66} & \underline{0.949} & \underline{0.041} & 71 & \underline{26.34} & \underline{0.932} & 0.052 & 30 & \underline{25.55} & \underline{0.921} & 0.063 & 33 \\
    Ours & \textbf{28.83} & \textbf{0.976} & \textbf{0.020} & \underline{73} & \textbf{30.18} & \textbf{0.964} & \textbf{0.025} & \underline{168} & \textbf{28.47} & \textbf{0.950} & \textbf{0.033} & \underline{112} & \textbf{28.21} & \textbf{0.943} & \textbf{0.037} & \underline{129} \\
    
    \bottomrule
  \end{tabular}
  }
  \label{tab:Comparison_sync}
\vspace{-3mm}
\end{table*}

\begin{table*}[t]
    \centering
    \caption{Quantitative results on our real-world dataset. Event-4DGS is an extension of Deformable-3DGS \cite{yang2024deformable} by incorporating events.}
    \vspace{-3mm}
    \setlength{\tabcolsep}{4pt}
    \renewcommand{\arraystretch}{1}
    \resizebox{1.0\linewidth}{!}{
      \begin{tabular}{c|cccc|cccc|cccc|cccc}
        \toprule
        \multirow{2}{*}{Method} & \multicolumn{4}{c|}{Excavator} & \multicolumn{4}{c|}{Jeep} & \multicolumn{4}{c|}{Flowers} & \multicolumn{4}{c}{Eagle} \\
        \cmidrule(lr){2-5} \cmidrule(lr){6-9} \cmidrule(lr){10-13} \cmidrule(lr){14-17}
        & PSNR$\uparrow$ & SSIM$\uparrow$ & LPIPS$\downarrow$ & FPS$\uparrow$ 
        & PSNR$\uparrow$ & SSIM$\uparrow$ & LPIPS$\downarrow$ & FPS$\uparrow$ 
        & PSNR$\uparrow$ & SSIM$\uparrow$ & LPIPS$\downarrow$ & FPS$\uparrow$
        & PSNR$\uparrow$ & SSIM$\uparrow$ & LPIPS$\downarrow$ & FPS$\uparrow$ \\
        \midrule
        4D-GS \cite{wu20244d} & 28.35 & 0.911 & 0.110 & \underline{115} & 28.34 & 0.878 & 0.093 & \underline{61} & 26.82 & 0.873 & 0.123 & 63 & 27.59 & 0.900 & 0.128 & \underline{105} \\
        Deformable-3DGS \cite{yang2024deformable} & 26.12 & 0.903 & 0.120 & 81 & 26.30 & 0.870 & 0.104 & 52 & 26.40 & 0.903 & \underline{0.079} & \underline{64} & 27.44 & \underline{0.903} & 0.125 & 70 \\
        Event-4DGS & \underline{29.67} & \underline{0.914} & \underline{0.092} & 57 & \underline{29.64} & \underline{0.901} & \underline{0.079} & 47 & \underline{27.53} & \underline{0.905} & 0.084 & 40 & \underline{29.08} & 0.896 & \underline{0.104} & 63 \\
        Ours & \textbf{31.28} & \textbf{0.925} & \textbf{0.070} & \textbf{179} & \textbf{30.41} & \textbf{0.905} & \textbf{0.068} & \textbf{89} & \textbf{28.57} & \textbf{0.913} & \textbf{0.069} & \textbf{149} & \textbf{31.29} & \textbf{0.918} & \textbf{0.074} & \textbf{192} \\
        \bottomrule
      \end{tabular}
    }
    \label{tab:Comparison_real}
    \vspace{-4mm}
\end{table*}

\noindent\textbf{Datasets.}
Current datasets for event-based dynamic scene reconstruction are highly limited, with only three synthetic and three real-world scenes \cite{ma2023deformable}, all unpublished. Notably, the only three publicly available real-world scenes \cite{ma2023deformable} were captured with a static camera, making novel view evaluation infeasible. To facilitate future research, we build the first event-inclusive 4D benchmark featuring 8 synthetic and 4 real-world dynamic scenes, encompassing diverse complexities, intricate structures, and rapid motions, thus enabling effective evaluation of dynamic reconstruction.

For synthetic scenes, we use Blender \cite{Blender2018} to generate one-second, 360° monocular camera rotations, producing thousands of continuous frames per scene. These high-temporal-resolution sequences are processed through ESIM \cite{rebecq2018esim} to generate events. For each sequence, we uniformly sample 30 frames (equivalent to 30 FPS) for training, and select intermediate frames as far apart as possible from the training frames for testing. Particularly, both ``Fan'' representing a typical \textit{high-speed} 4D scene, and ``Man'' featuring \textit{large object displacements}, present significant challenges.

\begin{figure}[t!]
  \centering
  \includegraphics[width=0.84\linewidth]{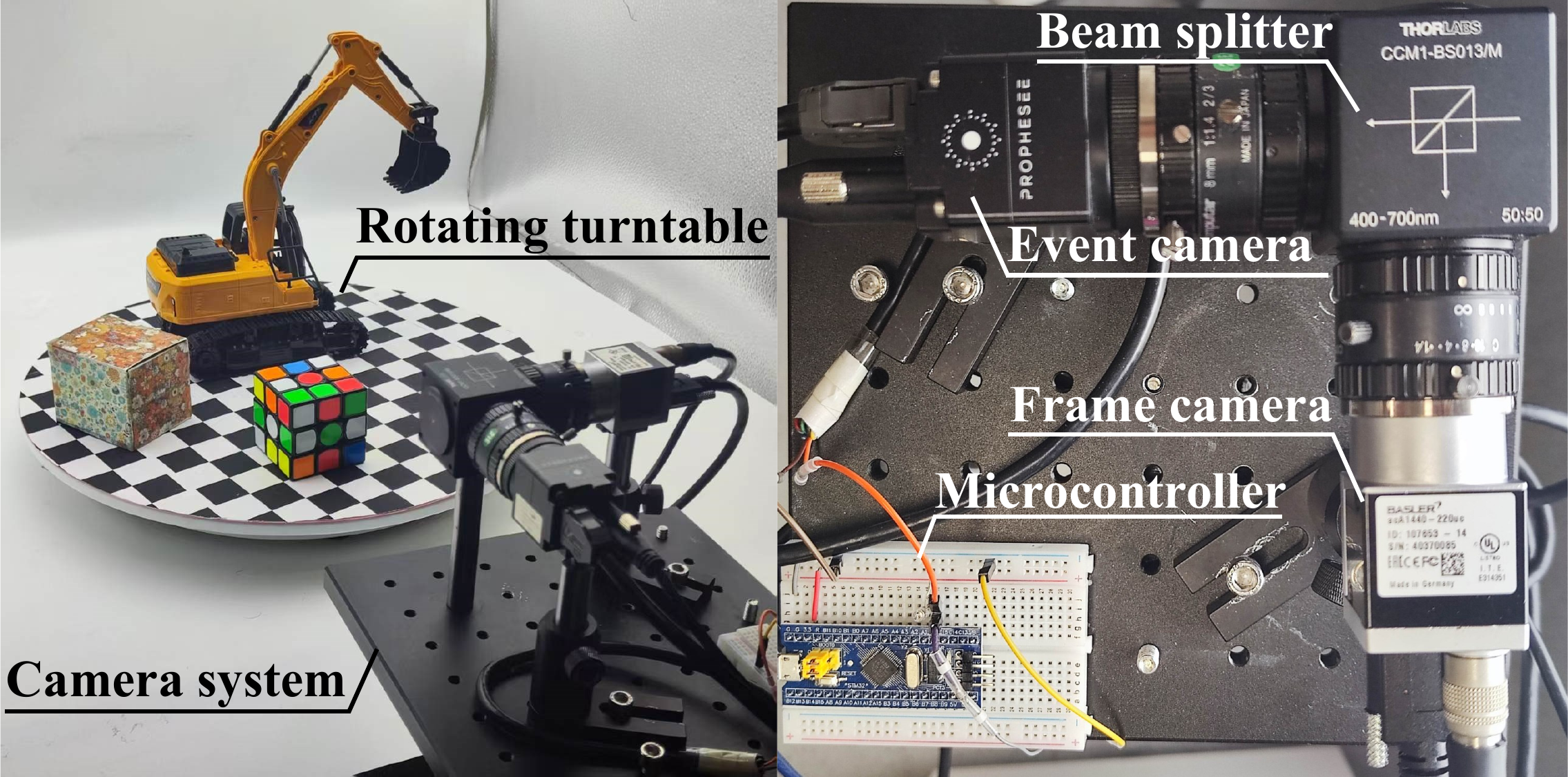}
  \vspace{-2mm}
  \caption{Real-world data acquisition setup (left) and our hybrid camera system (right).}
  \label{fig:system}
  \vspace{-4mm}
\end{figure}

For real-world scenes, as shown in \cref{fig:system}, we construct a hybrid camera system consisting of a beam splitter, an event camera (Prophesee Gen4), a frame camera (Basler ace), and a microcontroller (STM32) for outputting synchronization signals. Following \cite{rudnev2023eventnerf}, we keep the camera system static and place the objects on a motorized optical rotating turntable, which is equivalent to camera motion. Following prior work \cite{ma2023deformable}, we downsample the original high-FPS video for training and use intermediate frames for testing.

Our code, benchmark, and dataset creation pipeline will be publicly released, with \textit{more details provided in the supplementary materials.}

\noindent\textbf{Baselines.} For RGB-only settings, we benchmarked our method against the representative NeRF baselines K-Planes \cite{fridovich2023k} and TiNeuVox \cite{fang2022fast}, along with Gaussians baselines 3D-GS \cite{kerbl20233d}, 4D-GS \cite{wu20244d}, and Deformable-3DGS \cite{yang2024deformable}. For event-assisted settings, DE-NeRF \cite{ma2023deformable} is the only baseline; however, it could not be directly compared, as its code was still closed-source. DE-NeRF relies on NeRF's volume rendering techniques \cite{mildenhall2021nerf}, leading to predictably slow rendering speeds. Moreover, its reconstruction quality is also predictably limited due to the absence of threshold modeling for events. To provide a comparable baseline, we introduce Event-4DGS, an extension of Deformable-3DGS \cite{yang2024deformable} that incorporates the event rendering loss in \cref{eq:event_loss}.

\noindent\textbf{Metrics.} 
We evaluate rendering quality using PSNR, SSIM \cite{wang2004image}, and LPIPS \cite{zhang2018unreasonable} (based on AlexNet) and measure rendering speed in FPS on an NVIDIA RTX 3090 GPU.

\subsection{Comparisons}

\begin{figure*}[t!]
  \centering
  \includegraphics[width=0.9\linewidth]{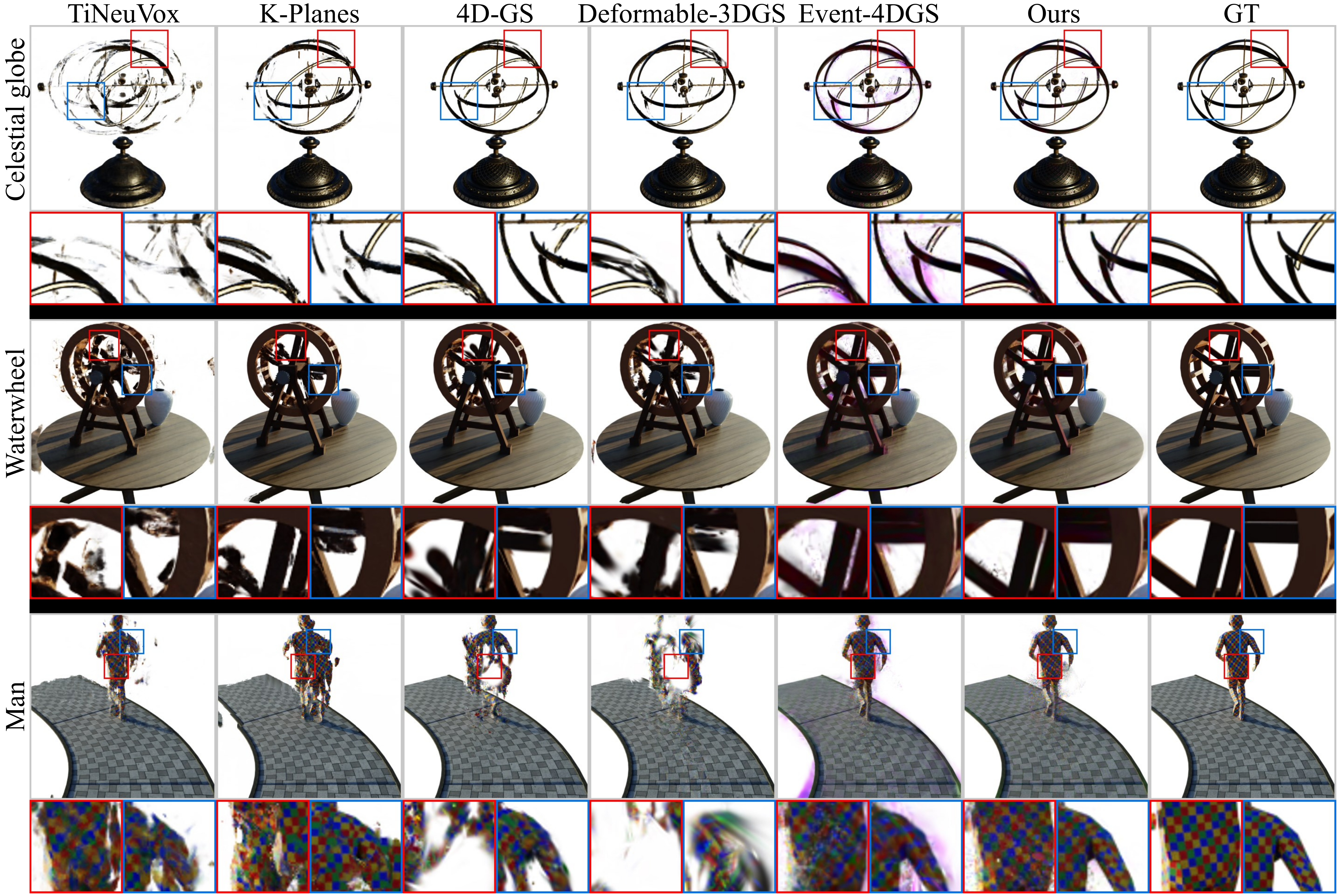}
  \vspace{-3mm}
  \caption{Qualitative comparisons on our synthetic dataset. \textbf{Please see the supplementary video for details.}}
  \label{fig:qual_comparison_Sync}
  \vspace{-3mm}
\end{figure*}

\begin{figure*}[t!]
  \centering
  \includegraphics[width=0.9\linewidth]{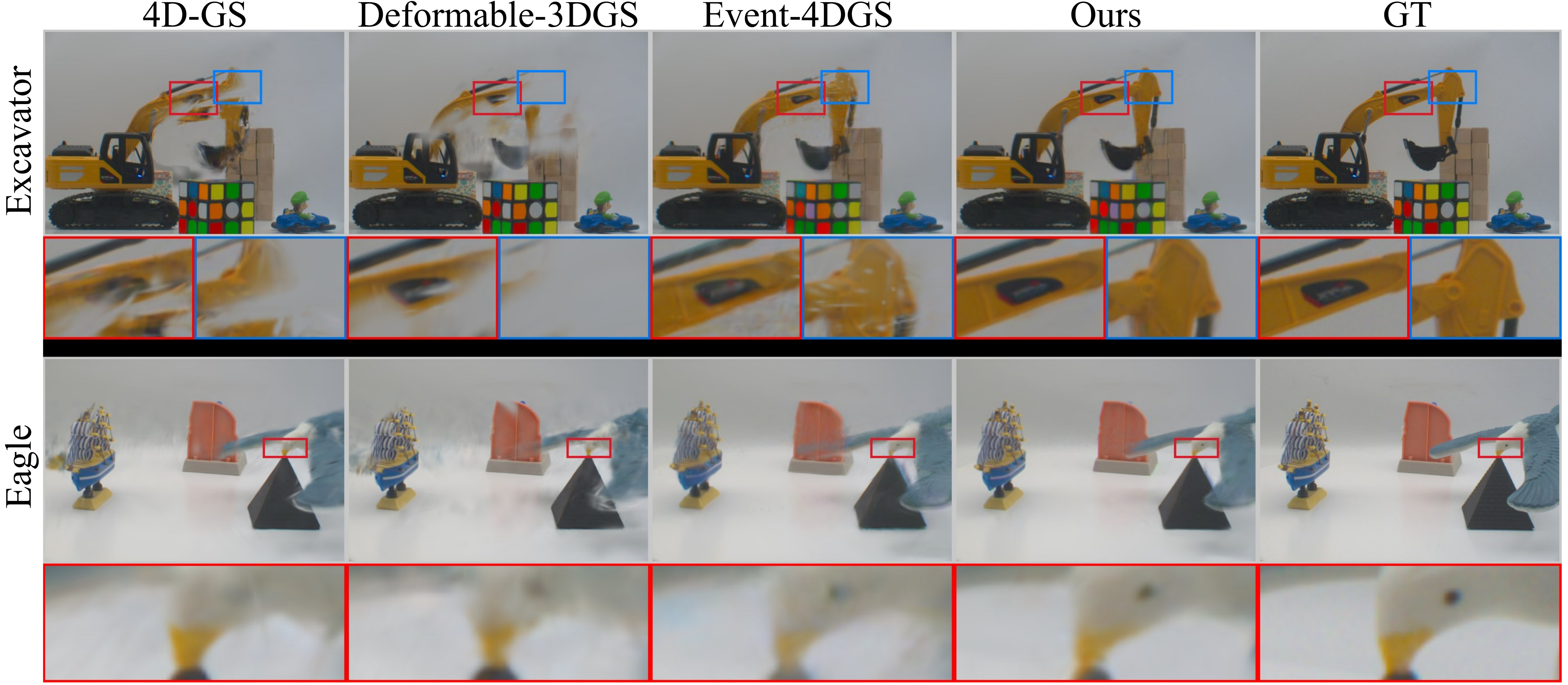}
  \vspace{-3mm}
  \caption{Qualitative comparisons on our real-world dataset. \textbf{Please see the supplementary video for details.}}
  \label{fig:qual_comparison_Real}
  \vspace{-4mm}
\end{figure*}

\noindent\textbf{Quantitative Results.}
We report the quantitative results of the comparison on the synthetic and real-world datasets in \cref{tab:Comparison_sync} and \cref{tab:Comparison_real}, respectively. Although 4D-GS \cite{wu20244d} and Deformable-3DGS \cite{yang2024deformable} achieve relatively higher FPS compared to NeRF baselines \cite{fridovich2023k, fang2022fast}, their reconstruction quality is limited by the sparsity of the RGB training frames. In contrast, Event-4DGS leverages the rich intermediate motion and viewpoint information provided by events, significantly outperforming other baselines in reconstruction quality, with an average PSNR improvement of 3.28 dB over Deformable-3DGS across all synthetic scenes. This notable improvement underscores the effectiveness of high-temporal-resolution event cameras for dynamic scene reconstruction. However, Event-4DGS still suffers from threshold variation, whereas our method with GTJM enables accurate threshold modeling and better event supervision, achieving an average PSNR improvement of 2.73 dB over Event-4DGS on synthetic datasets. Meanwhile, our method maintains exceptionally fast rendering speeds, averaging 1.71× faster than 4D-GS on synthetic datasets. In summary, our method enjoys both the highest rendering quality and exceptional rendering speed on both synthetic and real-world datasets.

\noindent\textbf{Qualitative Results.}
For a more visual assessment, we present qualitative results on the synthetic and real-world datasets in \cref{fig:qual_comparison_Sync} and \cref{fig:qual_comparison_Real}, respectively. These comparisons highlight the capability of our method to deliver high-fidelity dynamic scene modeling. Notably, our method effectively captures intricate motion details, while other baselines exhibit structural deficiencies and distortions.

\begin{figure}[t!]
  \centering
  \includegraphics[width=0.95\linewidth]{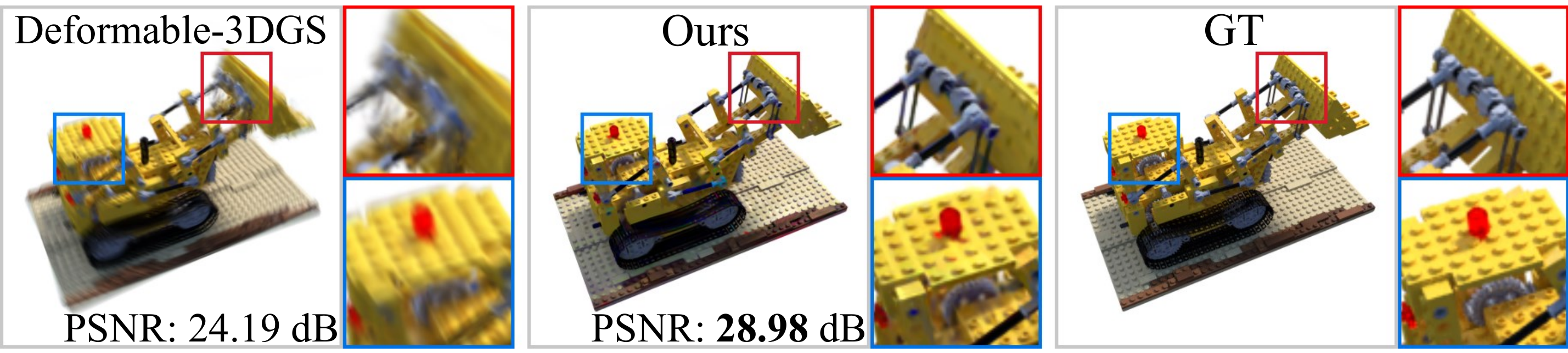}
  \vspace{-3mm}
  \caption{Extended comparisons on the dynamic blurry scene.}
  \label{fig:BlurryLego}
  \vspace{-4mm}
\end{figure}

\noindent\textbf{Dynamic Blurry Scene Comparisons.}
Motion blur is another common challenge in dynamic scenes. To address this, we extend both baselines and our method with blur loss and EDI from \cite{yu2024evagaussians}, and build blurry scenes for evaluations. \cref{fig:BlurryLego} shows that, by leveraging events' deblurring advantage, our method outperforms Deformable-3DGS by 4.79 dB in PSNR, achieving the best results. For detailed methods and quantitative results, see the supplementary material.

\begin{table}[t]
    \centering
    \caption{Ablation studies on synthetic dataset. For real-world ablation studies, please refer to the supplementary materials.}
    \vspace{-3mm}
    \setlength{\tabcolsep}{1pt}
    \renewcommand{\arraystretch}{1}
    \resizebox{1.0\linewidth}{!}{
      \begin{tabular}{c|cccc}
        \toprule
        Method & PSNR$\uparrow$ & SSIM$\uparrow$ & LPIPS$\downarrow$ & FPS$\uparrow$ \\
        \midrule
        w/o GTJM & 29.39 & 0.956 & 0.034 & 153 \\
        w/o Joint Optimization in GTJM & 30.87 & 0.963 & 0.026 & 152 \\
        w/o DSD & 30.78 & 0.961 & 0.026 & 57 \\
        w/o Buffer-based Soft Decomposition & 31.02 & 0.963 & 0.025 & 138 \\
        Full & \textbf{31.56} & \textbf{0.966} & \textbf{0.022} & \textbf{156} \\
        \bottomrule
      \end{tabular}
    }
    \label{tab:Ablation_sync}
    \vspace{-6mm}
\end{table}

\subsection{Ablation Study}
\label{sec:ablation}

\begin{figure}[t!]
  \centering
  \includegraphics[width=0.95\linewidth]{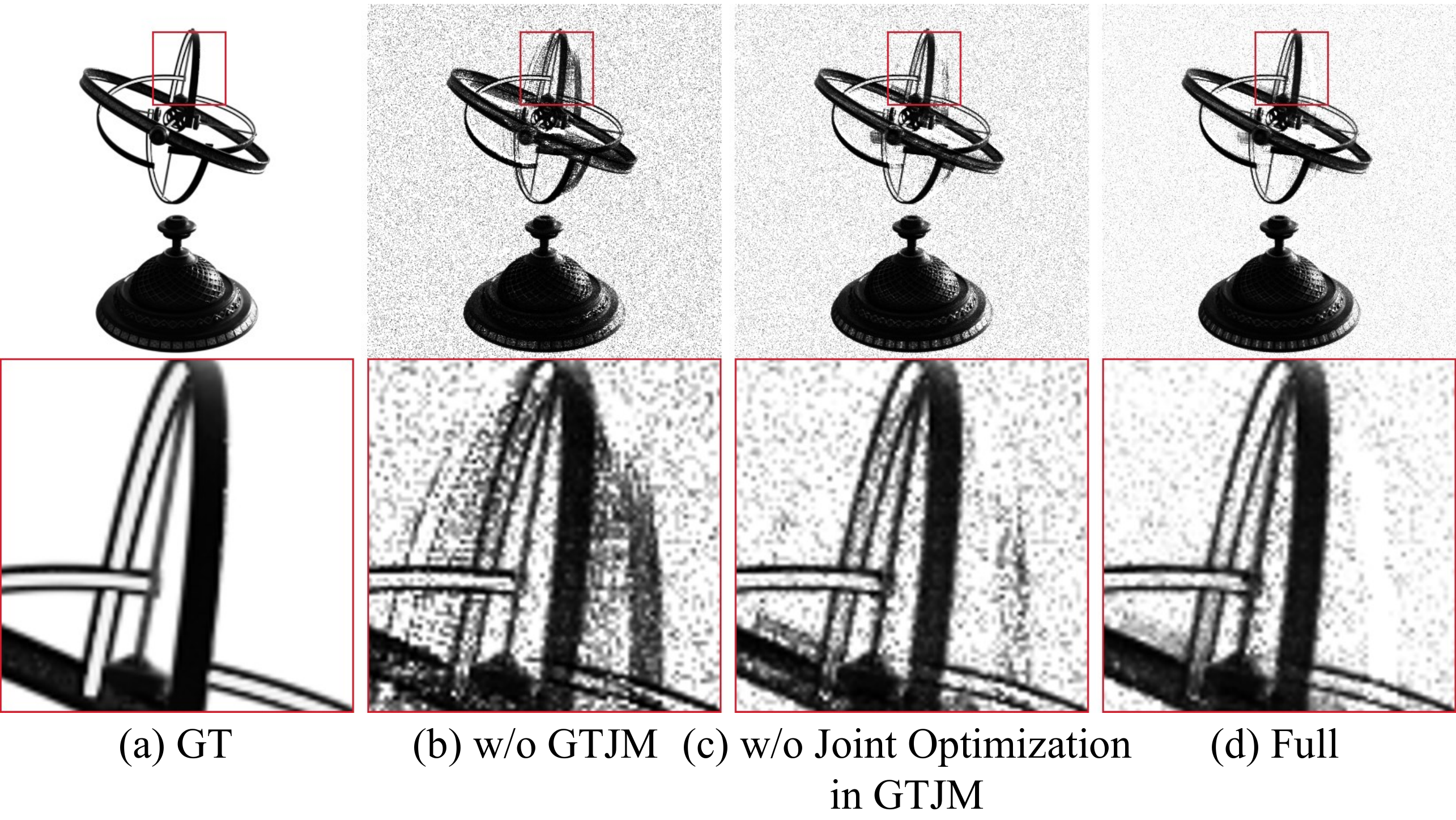}
  \vspace{-3mm}
  \caption{The effect of GS-threshold joint modeling strategy, which eliminates event artifacts caused by threshold variations.}
  \label{fig:Thres_modeling_vis}
  \vspace{-5mm}
\end{figure}

\noindent\textbf{GS-threshold Joint Modeling.}
Using a constant threshold fails to properly neutralize opposing polarity events during accumulation, resulting in motion trajectory artifacts as shown in \cref{fig:Thres_modeling_vis} (b). These artifacts, when used for Gaussian supervision, produce undesirable purple haze in rendered outputs, such as the Event-4DGS results in \cref{fig:qual_comparison_Sync}. Our RGB-assisted threshold estimation significantly reduces these artifacts (\cref{fig:Thres_modeling_vis} (c)), while subsequent joint threshold and GS optimization effectively eliminates remaining distortions (\cref{fig:Thres_modeling_vis} (d)). As demonstrated in \cref{tab:Ablation_sync}, this improved event supervision yields a 2.17 dB average PSNR improvement across all scenes, validating our GTJM strategy's effectiveness in handling threshold variations.

\begin{figure}[t!]
  \centering
  \includegraphics[width=0.9\linewidth]{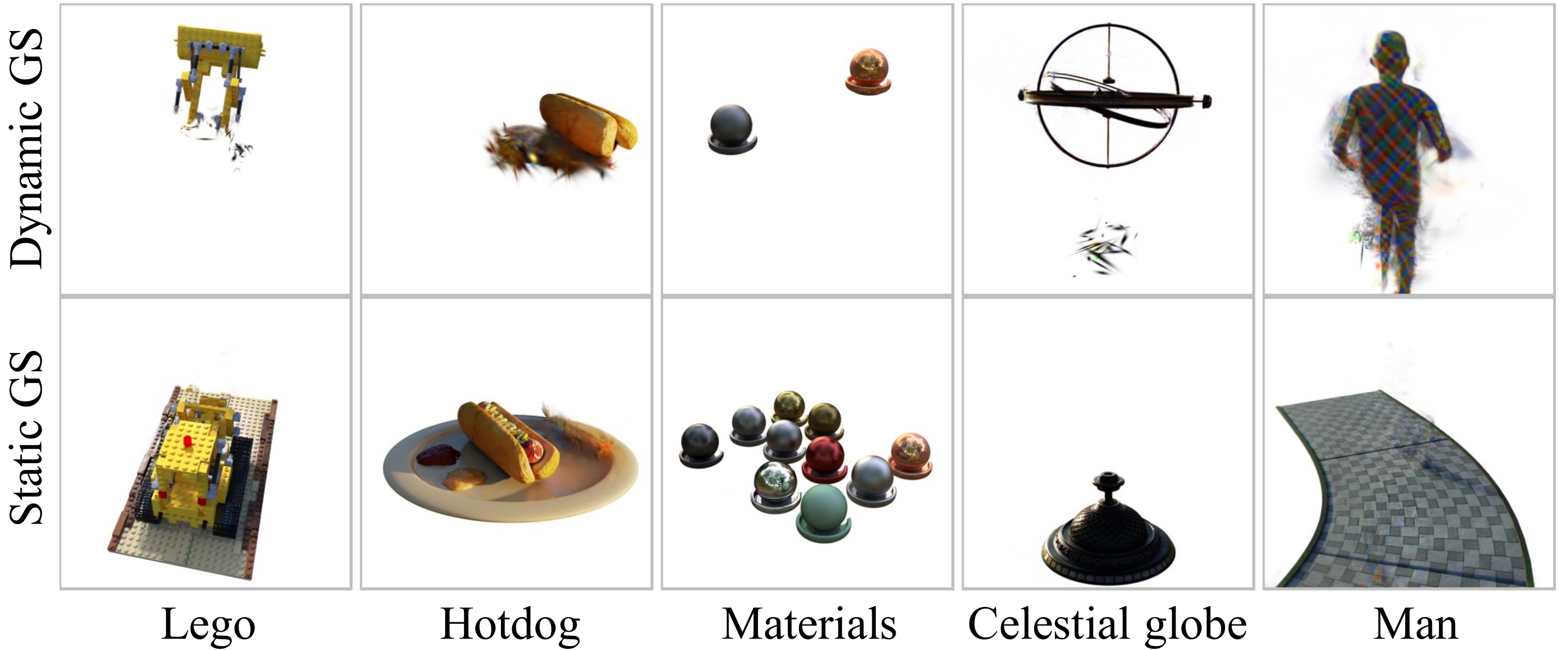}
  \vspace{-2mm}
  \caption{Rendering results of dynamic and static Gaussians separated by our dynamic-static decomposition strategy.}
  \label{fig:decomp_show}
  \vspace{-4mm}
\end{figure}

\noindent\textbf{Dynamic-static Decomposition.}
Our DSD method successfully identifies dynamic regions of varying sizes and geometries, as demonstrated in \cref{fig:decomp_show}. Modeling the entire scene with dynamic Gaussians without DSD misallocates deformation field capacity to static regions, compromising dynamic region reconstruction quality as shown in \cref{fig:wo_decomp_comparison}. Quantitative results in \cref{tab:Ablation_sync} demonstrate that using DSD improves the average PSNR by 0.78 dB and accelerates the rendering speed to 2.74 times the original FPS. This underscores DSD's crucial role in achieving both high-fidelity dynamic scene reconstruction and efficient rendering.

\begin{figure}[t!]
  \centering
  \includegraphics[width=0.85\linewidth]{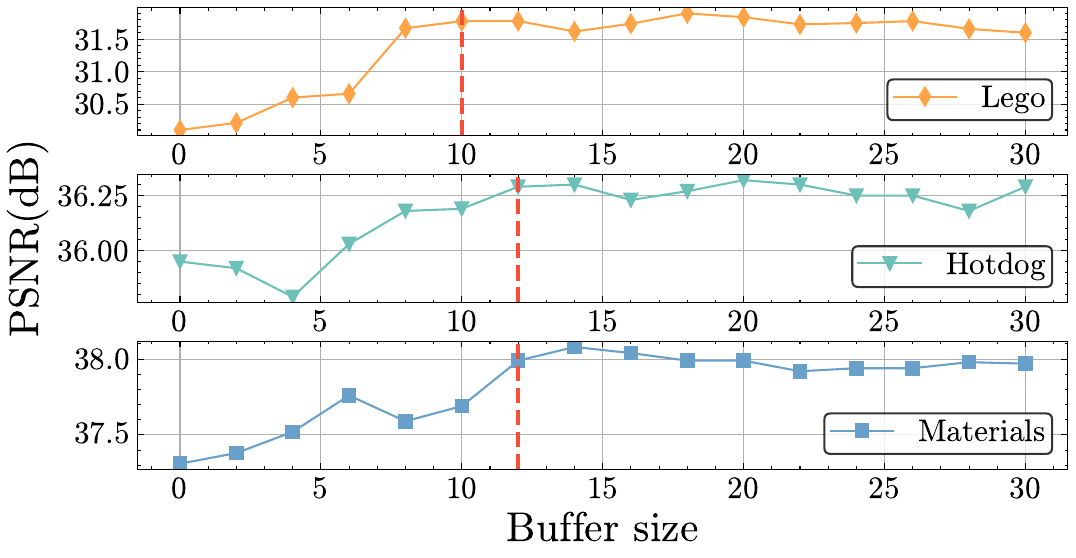}
  \vspace{-4mm}
  \caption{Sensitivity analysis on buffer size ($r_{2}-r_{1}$).}
  \label{fig:buffer_size}
  \vspace{-5mm}
\end{figure}

\noindent\textbf{Buffer-based Soft Decomposition.}
Our buffer-based soft decomposition enables adaptive optimization of decomposition boundaries, yielding a 0.54 dB improvement in average PSNR (\cref{tab:Ablation_sync}). Sensitivity analysis reveals that reconstruction quality stabilizes when buffer size ($r_{2}-r_{1}$) exceeds approximately 12 basic units (normalized by average inter-Gaussian distance to account for scene variations), as shown in \cref{fig:buffer_size}. This stability demonstrates the robustness of our DSD method through adaptive boundary search, highlighting the effectiveness of the buffer-based strategy.

\section{Conclusion}
In this paper, we present an event-boosted deformable 3D Gaussian framework for high-quality dynamic scene reconstruction. Our GS-threshold joint modeling effectively addresses threshold variation challenges, enabling reliable event supervision. The proposed dynamic-static decomposition method enhances both rendering efficiency and reconstruction quality through optimized resource allocation between static and dynamic regions.

{
    \small
    \bibliographystyle{ieeenat_fullname}
    \bibliography{main}
}

\end{document}


\maketitle


This supplementary material is organized as follows:

\cref{sec:supp_quan_results} presents additional quantitative results.

\cref{sec:supp_dataset} presents further dataset details.

\cref{sec:supp_qual_results} presents additional qualitative results.

\cref{sec:supp_implementation} presents implementation details.


\section{Additional Quantitative Results}
\label{sec:supp_quan_results}
\subsection{Additional Ablation Studies}

\begin{table*}[h!]
  \caption{Per-scene ablation studies on synthetic dataset.}
  \centering
    \setlength{\tabcolsep}{5pt}
  \renewcommand{\arraystretch}{1}
  \resizebox{1.0\linewidth}{!}{
  \begin{tabular}{c|cccc|cccc|cccc|cccc}
    \toprule

    \multirow{2}{*}{\centering Method} & \multicolumn{4}{c|}{Lego} & \multicolumn{4}{c|}{Hotdog} & \multicolumn{4}{c|}{Materials} & \multicolumn{4}{c}{Music box} \\
    
    \cmidrule(lr){2-5} \cmidrule(lr){6-9} \cmidrule(lr){10-13} \cmidrule(lr){14-17}
    
    & PSNR$\uparrow$ & SSIM$\uparrow$ & LPIPS$\downarrow$ & FPS$\uparrow$ & PSNR$\uparrow$ & SSIM$\uparrow$ & LPIPS$\downarrow$ & FPS$\uparrow$ & PSNR$\uparrow$ & SSIM$\uparrow$ & LPIPS$\downarrow$ & FPS$\uparrow$ & PSNR$\uparrow$ & SSIM$\uparrow$ & LPIPS$\downarrow$ & FPS$\uparrow$ \\

    \cmidrule(lr){1-1} \cmidrule(lr){2-5} \cmidrule(lr){6-9} \cmidrule(lr){10-13} \cmidrule(lr){14-17}

    w/o GTJM & 29.49 & 0.955 & 0.028 & 177 & 34.72 & 0.970 & 0.017 & \textbf{261} & 35.72 & 0.990 & 0.006 & \textbf{243} & 28.73 & 0.952 & 0.042 & 79 \\
    w/o Joint Optimization in GTJM & 30.92 & 0.962 & 0.021 & 185 & 35.67 & 0.973 & 0.014 & 238 & 37.27 & 0.992 & 0.004 & 239 & 30.15 & 0.959 & 0.033 & 87 \\
    w/o DSD & 29.62 & 0.954 & 0.028 & 58 & 35.69 & 0.972 & 0.014 & 104 & 37.70 & 0.992 & 0.004 & 76 & 30.29 & 0.959 & 0.032 & 43 \\
    w/o Buffer-based Soft Decomposition & 30.27 & 0.959 & 0.025 & 168 & 35.95 & 0.973 & 0.015 & 216 & 37.31 & 0.991 & 0.004 & 207 & 30.77 & \textbf{0.963} & 0.030 & 82 \\
    Full & \textbf{31.85} & \textbf{0.967} & \textbf{0.018} & \textbf{189} & \textbf{36.15} & \textbf{0.974} & \textbf{0.013} & 241 & \textbf{38.02} & \textbf{0.993} & \textbf{0.003} & 240 & \textbf{30.78} & \textbf{0.963} & \textbf{0.029} & \textbf{92} \\

    \midrule
    
    \multirow{2}{*}{\centering Method} & \multicolumn{4}{c|}{Celestial globe} & \multicolumn{4}{c|}{Fan} & \multicolumn{4}{c|}{Water wheel} & \multicolumn{4}{c}{Man} \\
    
    \cmidrule(lr){2-5} \cmidrule(lr){6-9} \cmidrule(lr){10-13} \cmidrule(lr){14-17}
    
    & PSNR$\uparrow$ & SSIM$\uparrow$ & LPIPS$\downarrow$ & FPS$\uparrow$ & PSNR$\uparrow$ & SSIM$\uparrow$ & LPIPS$\downarrow$ & FPS$\uparrow$ & PSNR$\uparrow$ & SSIM$\uparrow$ & LPIPS$\downarrow$ & FPS$\uparrow$ & PSNR$\uparrow$ & SSIM$\uparrow$ & LPIPS$\downarrow$ & FPS$\uparrow$ \\

    \cmidrule(lr){1-1} \cmidrule(lr){2-5} \cmidrule(lr){6-9} \cmidrule(lr){10-13} \cmidrule(lr){14-17}

    w/o GTJM & 25.06 & 0.955 & 0.038 & 67 & 28.16 & 0.955 & 0.035 & \textbf{169} & 26.57 & 0.932 & 0.051 & \textbf{113} & 26.67 & 0.937 & 0.053 & 118 \\
    w/o Joint Optimization in GTJM & 27.49 & 0.970 & 0.025 & 74 & 29.54 & 0.961 & 0.028 & 165 & 28.14 & 0.945 & 0.038 & 110 & 27.76 & 0.940 & 0.041 & 120 \\
    w/o DSD & 27.83 & 0.971 & 0.024 & 37 & 29.73 & 0.959 & 0.029 & 77 & 27.86 & 0.948 & 0.034 & 31 & 27.54 & 0.935 & 0.040 & 31 \\
    w/o Buffer-based Soft Decomposition & 28.16 & 0.973 & 0.022 & \textbf{78} & 29.65 & 0.959 & 0.029 & 153 & 28.36 & 0.949 & \textbf{0.033} & 90 & 27.67 & 0.938 & 0.038 & 106 \\
    Full & \textbf{28.83} & \textbf{0.976} & \textbf{0.020}& 73 & \textbf{30.18} & \textbf{0.964} & \textbf{0.025} & 168 & \textbf{28.47} & \textbf{0.950} & \textbf{0.033} & 112 & \textbf{28.21} & \textbf{0.943} & \textbf{0.037} & \textbf{129} \\
    
    \bottomrule
  \end{tabular}
  }
  \label{tab:Ablation_Sync}
\end{table*}

\begin{table*}[h!]
  \caption{Per-scene ablation studies on real-world dataset.}
    \centering
    \setlength{\tabcolsep}{5pt}
    \renewcommand{\arraystretch}{1}
    \resizebox{1.0\linewidth}{!}{
      \begin{tabular}{c|cccc|cccc|cccc|cccc}
        \toprule
        \multirow{2}{*}{Method} & \multicolumn{4}{c|}{Excavator} & \multicolumn{4}{c|}{Jeep} & \multicolumn{4}{c|}{Flowers} & \multicolumn{4}{c}{Eagle} \\
        \cmidrule(lr){2-5} \cmidrule(lr){6-9} \cmidrule(lr){10-13} \cmidrule(lr){14-17}
        & PSNR$\uparrow$ & SSIM$\uparrow$ & LPIPS$\downarrow$ & FPS$\uparrow$ 
        & PSNR$\uparrow$ & SSIM$\uparrow$ & LPIPS$\downarrow$ & FPS$\uparrow$ 
        & PSNR$\uparrow$ & SSIM$\uparrow$ & LPIPS$\downarrow$ & FPS$\uparrow$ 
        & PSNR$\uparrow$ & SSIM$\uparrow$ & LPIPS$\downarrow$ & FPS$\uparrow$ \\
        
        \midrule
        
        w/o GTJM & 30.30 & 0.918 & 0.083 & 169 & 29.64 & 0.900 & 0.081 & 79 & 28.00 & 0.910 & 0.081 & 149 & 30.80 & 0.914 & 0.084 & 178 \\
        w/o Joint Optimization in GTJM & 31.14 & 0.923 & 0.073 & 169 & 30.25 & 0.904 & 0.072 & \textbf{91} & 28.44 & 0.912 & 0.071 & 140 & 31.14 & 0.916 & 0.076 & 185 \\
        w/o DSD & 30.14 & 0.913 & 0.085 & 60 & 30.13 & 0.904 & 0.072 & 42 & 27.76 & 0.907 & 0.078 & 44 & 29.36 & 0.888 & 0.082 & 57 \\
        w/o Buffer-based Soft Decomposition & 30.67 & 0.920 & 0.077 & 166 & 30.25 & 0.903 & 0.072 & 84 & 28.49 & \textbf{0.915} & 0.071 & \textbf{158} & 31.05 & 0.915 & 0.077 & \textbf{193} \\
        Full & \textbf{31.28} & \textbf{0.925} & \textbf{0.070} & \textbf{179} & \textbf{30.41} & \textbf{0.905} & \textbf{0.068} & 89 & \textbf{28.57} & 0.913 & \textbf{0.069} & 149 & \textbf{31.29} & \textbf{0.918} & \textbf{0.074} & 192 \\

        \bottomrule
      \end{tabular}
    }
    \label{tab:Ablation_Real}
\end{table*}

\noindent\textbf{Per-scene Ablation Study.}
As shown in \cref{tab:Ablation_Sync} and \cref{tab:Ablation_Real}, we present the per-scene ablation results of our method across both synthetic and real-world scenes. The results provide compelling evidence that each proposed component contributes to the overall performance. 

Notably, the optimal FPS does not show a clear pattern, as ablating components indirectly affects the number of Gaussians, leading to \textit{acceptable fluctuations} in FPS. However, what truly stands out is that \textit{ablating DSD significantly reduces FPS}, highlighting the crucial role of DSD in accelerating rendering.
In particular, as demonstrated in \cref{tab:Ablation_Real}, using GTJM leads to an average PSNR improvement of 0.70 dB, indicating that \textit{our method effectively models the threshold variations of \underline{real event cameras}}. Meanwhile, in real-world scenes, using DSD also brings an average PSNR improvement of 1.04 dB and a 2.96× speedup in rendering. In summary, \textit{the key components of our method, GTJM and DSD, remain effective in \underline{real-world} scenes}, enabling our method to achieve high-fidelity reconstruction quality and exceptional rendering speed.

\begin{figure}[h!]
  \centering
  \includegraphics[width=0.5\linewidth]{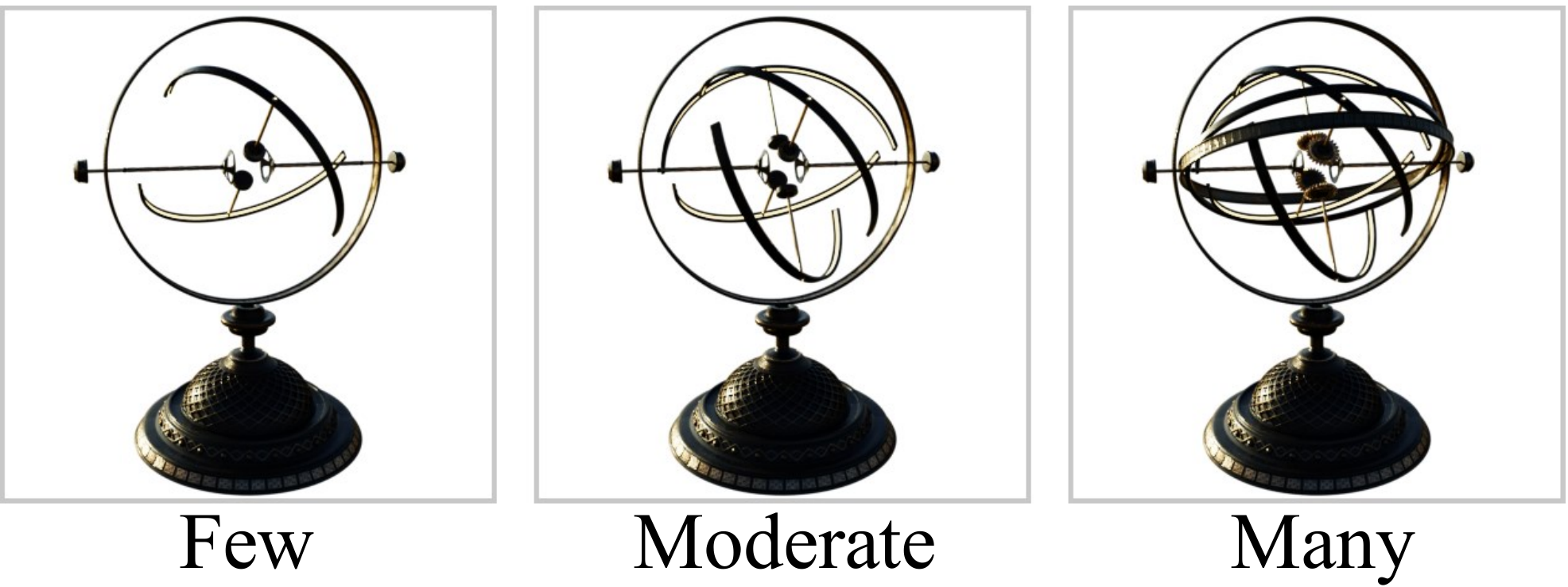}
  \caption{Scenes with varying numbers of dynamic components.}
  \label{fig:Supp_mode_showing}
\end{figure}

\noindent\textbf{The Effect of Motion Complexity on Dynamic-static Decomposition Strategy.} We measure  motion complexity from two aspects: dynamic component count and motion speed. For dynamic component count, we extend the original scene ``Moderate'' by adding or removing dynamic components, creating two variant scenes, ``Many'' and ``Few'' (see \cref{fig:Supp_mode_showing}). For motion speed, we adjust the animation speed in Blender \cite{Blender2018} to create two variant scenes, ``Slow'' (0.5x the original speed) and ``Fast'' (2.0x the original speed).

As shown in \cref{tab:motion_complexity}, applying the DSD strategy enables the deformation field to focus on the dynamic regions, resulting in a significant average PSNR improvement of 0.99 dB across all scenes. More importantly, we observe that the gains from the DSD strategy increase as motion complexity grows. This highlights the potential of our method in reconstructing complex dynamic scenes.

\begin{table}[h]
  \caption{The effect of motion complexity on dynamic-static decomposition strategy.}
  \centering
    \setlength{\tabcolsep}{17pt}
  \renewcommand{\arraystretch}{1}
  \resizebox{0.85\linewidth}{!}{
  \begin{tabular}{c|ccc|ccc}
    \toprule
    
    \multirow{2}{*}{\centering Method} & \multicolumn{3}{c|}{Dynamic component count} & \multicolumn{3}{c}{Motion speed} \\
    \cmidrule(lr){2-4} \cmidrule(lr){5-7}
    
    & Few & Moderate & Many & Slow & Moderate & Fast \\
    \midrule
    
    w/o DSD & 30.50 & 27.83 & 26.79 & 28.92 & 27.83 & 24.96 \\ 
    Full & \textbf{31.25} & \textbf{28.83} & \textbf{28.09} & \textbf{29.60} & \textbf{28.83} & \textbf{26.16} \\

    \midrule
    $\bigtriangleup$ & 0.75 & 1.00 & 1.30 & 0.68 & 1.00 & 1.20 \\ 

    \bottomrule
  \end{tabular}
  }
  \label{tab:motion_complexity}
\end{table}

\begin{figure}[h!]
  \centering
  \includegraphics[width=0.6\linewidth]{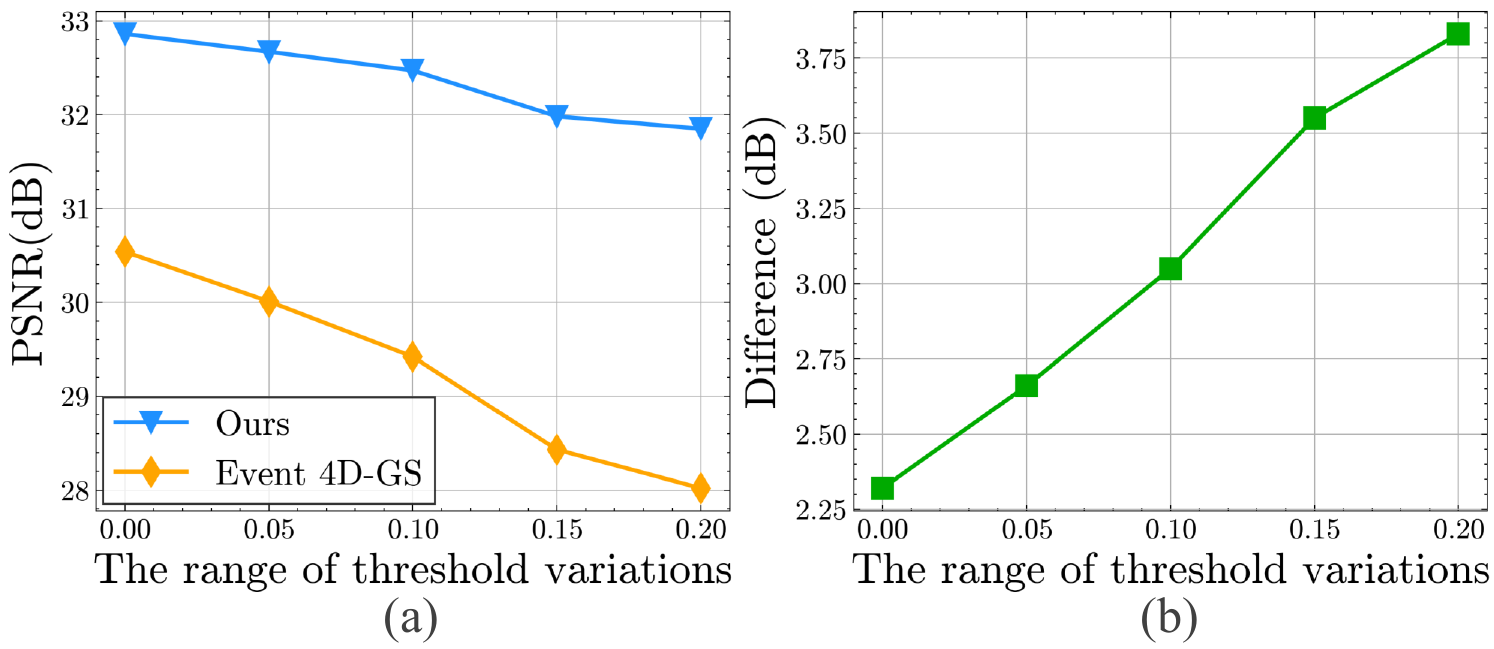}
  \caption{(a) The effect of different ranges of threshold variations on Event-4DGS and our method.
  (b) PSNR differences between Event-4DGS and our method across different ranges of threshold variations.
  }
  \label{fig:Supp_C_vary}
\end{figure}

\noindent\textbf{The Effect of Varying Degrees of Threshold Variations.} The greater the range of threshold variations, the more intense the degree of threshold variations becomes. As shown in \cref{fig:Supp_C_vary} (a), the performance of both our method and the baseline Event-4DGS declines under intense threshold variations, with Event-4DGS experiencing a more significant decline. \cref{fig:Supp_C_vary} (b) shows that the PSNR difference between our method and Event-4DGS widens as the threshold variations become more intense. For instance, at a variation range of 0.05, our method achieves a 2.66 dB PSNR improvement over Event-4DGS. When the range increases to 0.20, this improvement grows to 3.83 dB. These results demonstrate that our proposed threshold modeling method becomes increasingly beneficial as threshold variations become more intense.

\begin{figure}[h!]
  \centering
  \includegraphics[width=0.6\linewidth]{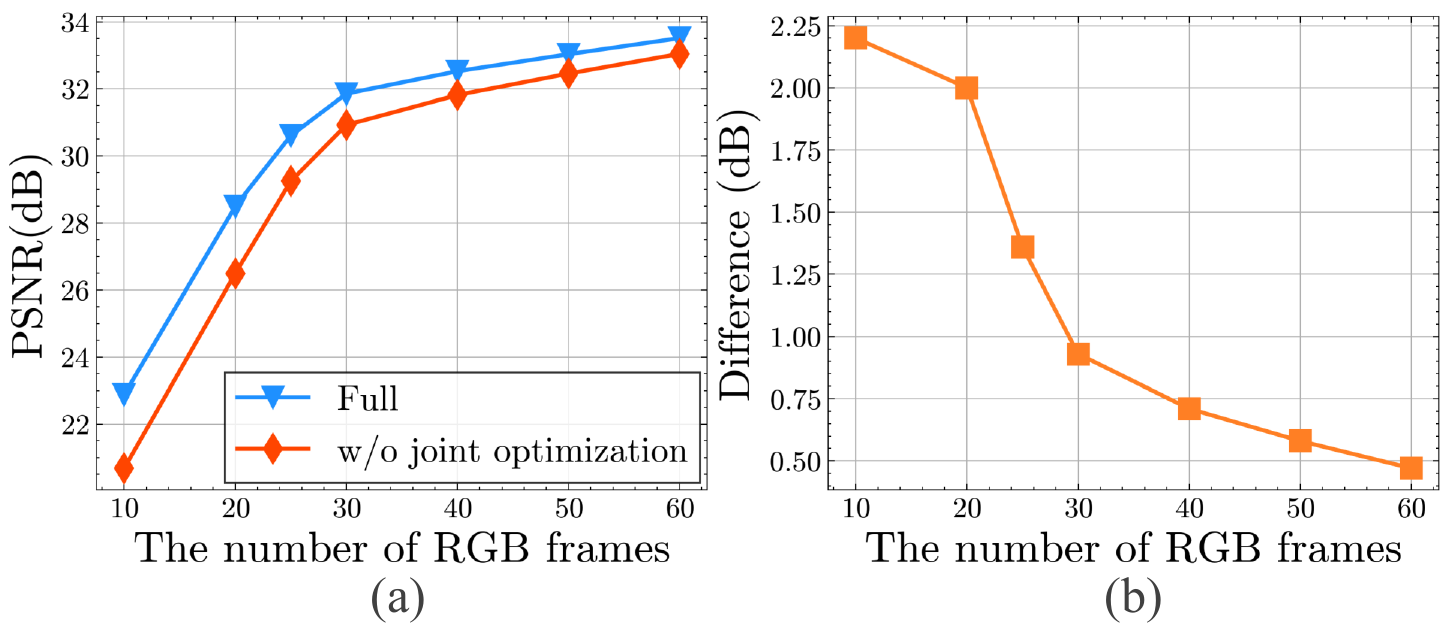}
  \caption{(a) The effect of different number of RGB frames on our method and its variant (without joint optimization in GTJM).
  (b) PSNR differences between our method and its variant across different numbers of RGB frames.
  }
  \label{fig:Supp_RGB_num_vary}
\end{figure}

\noindent\textbf{The Effect of Different Number of RGB Frames.} In \cref{fig:Supp_RGB_num_vary} (a), we demonstrate the effect of different numbers of RGB frames on our method and its variant (without joint optimization in GTJM). Denser RGB frames provide stronger geometric supervision, naturally benefiting both our method and the variant. More importantly, as shown in \cref{fig:Supp_RGB_num_vary} (b), the PSNR difference between our method and the variant increases as the RGB frames become sparser. Notably, in the extremely sparse case with only 10 RGB frames, applying joint optimization in GTJM delivers a significant 2.20 dB improvement. These results support the motivation behind \textit{GS-boosted threshold refinement} in Section 3.2 of the main paper, which leverages 3D-GS rendered intermediate frames as additional pseudo supervision to enhance the limited supervision from sparse RGB frames.

\noindent\textbf{Evaluation on Longer Videos.} 
Considering that a one-second motion may not be sufficient, we further evaluate our method on the ``Lego" scene with longer capture inputs. As shown in \cref{tab:longer_videos}, as the number of captured inputs increases, the reconstruction quality of our method improves, demonstrating its robustness on longer videos.

\begin{table}[h]
  \caption{Evaluation on longer videos.}
  \centering
    \setlength{\tabcolsep}{10pt}
  \renewcommand{\arraystretch}{1}
  \resizebox{0.6\linewidth}{!}{
  \begin{tabular}{c|cccccc}
    \toprule

    Time(s) & 1.0 & 2.0 & 3.0 & 4.0 & 5.0 & 6.0 \\
    \midrule
    PSNR & 31.85 & 33.47 & 34.18 & 34.69 & 35.04 & 35.18 \\
    
    \bottomrule
  \end{tabular}
  }
  \label{tab:longer_videos}
\end{table}

\subsection{Dynamic Blurry Scene Reconstruction}
\label{sec:dynamic_blurry}

In dynamic scene reconstruction, severely motion-blurred training frames pose a significant challenge. Such frames exhibit inaccurate multi-view geometric correspondences, making it difficult for 3D-GS to represent the 3D scene. To address this challenge, we extend both the baselines and our approach by explicitly modeling the motion blur process.

First, we replace the original RGB rendering loss Eq.(5) with the blur reconstruction loss from \cite{yu2024evagaussians} to model the motion blur process. Specifically, we render $n$ images $\left \{ \hat{I}_{i} \right \} _{i=1}^{n}$ from 3D-GS along the camera trajectory during the exposure period and approximate motion blur formation via discrete averaging,

\begin{equation}
\hat{B}=\frac{1}{n}\sum_{i=1}^{n} \hat{I}_{i}.
\end{equation}

We then minimize the photometric error between the simulated blurry image $\hat{B}$ and the ground truth blurry image $B$ as follows,

\begin{equation}
\mathcal{L}_{blur}=(1-\lambda_{s})\left \| \hat{B}-B\right \|_{1}+\lambda_{s}\mathcal{L}_{D-SSIM}(\hat{B},B).
\end{equation}

Second, for our method and Event-4DGS, we integrate the Event-based Double Integral (EDI) model \cite{pan2019bringing}, which derives sharp latent frames from motion-blurred inputs. The resulting sharp frames serve as $I(t)$ in Eq.(4) of the main paper, enabling the computation of the event rendering loss. Additionally, our method's GTJM and DSD components are also built upon these sharp frames, allowing them to function as intended.


\begin{table}[h]
\centering

\makebox[\textwidth]{ 
\begin{minipage}{0.4\linewidth}
    \caption{Comparison on ``Blurry Lego''.}
    \vspace{-3mm}
    \centering
    \setlength{\tabcolsep}{1.0pt}
    \renewcommand{\arraystretch}{1}
    \resizebox{1.0\linewidth}{!}{
    \begin{tabular}{c|ccc}
    \toprule
        Method & PSNR$\uparrow$ & SSIM$\uparrow$ & LPIPS$\downarrow$ \\
    \midrule
        Deformable-3DGS & 24.19 & 0.866 & 0.156 \\
        Event-4DGS & \underline{27.41} & \underline{0.921} & \underline{0.054} \\
        Ours & \textbf{28.98} & \textbf{0.934} & \textbf{0.047} \\
    \bottomrule
    \end{tabular}
    }
    \label{tab:Blurry_Lego_Comp}
\end{minipage}
\hspace{5mm} 
\begin{minipage}{0.35\linewidth}
    \caption{Ablation studies on ``Blurry Lego''.}
    \vspace{-3mm}
    \centering
    \setlength{\tabcolsep}{1.0pt}
    \renewcommand{\arraystretch}{1}
    \resizebox{1.0\linewidth}{!}{
    \begin{tabular}{c|ccc}
    \toprule
        Method & PSNR$\uparrow$ & SSIM$\uparrow$ & LPIPS$\downarrow$ \\
    \midrule
        w/o GTJM & 28.25 & 0.927 & 0.050 \\
        w/o DSD & 28.07 & 0.928 & 0.051 \\
        Full & \textbf{28.98} & \textbf{0.934} & \textbf{0.047}  \\
    \bottomrule
    \end{tabular}
    }
    \label{tab:Blurry_Lego_Abl}
\end{minipage}
}
\end{table}

For evaluation, we construct a dynamic blurry scene, ``Blurry Lego'', by averaging 32 consecutive latent frames. In \cref{tab:Blurry_Lego_Comp}, we report the quantitative comparison results on ``Blurry Lego'' corresponding to Fig.9 of the main paper. The results demonstrate that by leveraging the deblurring advantage of events, our method still achieves the best performance. Meanwhile, as shown in \cref{tab:Blurry_Lego_Abl}, our key components, GTJM and DSD, remain robust to motion blur.

\section{More Details on Dataset}
\label{sec:supp_dataset}

\subsection{More Details on Synthetic Dataset}

\begin{figure*}[h!]
  \centering
  \includegraphics[width=1.0\linewidth]{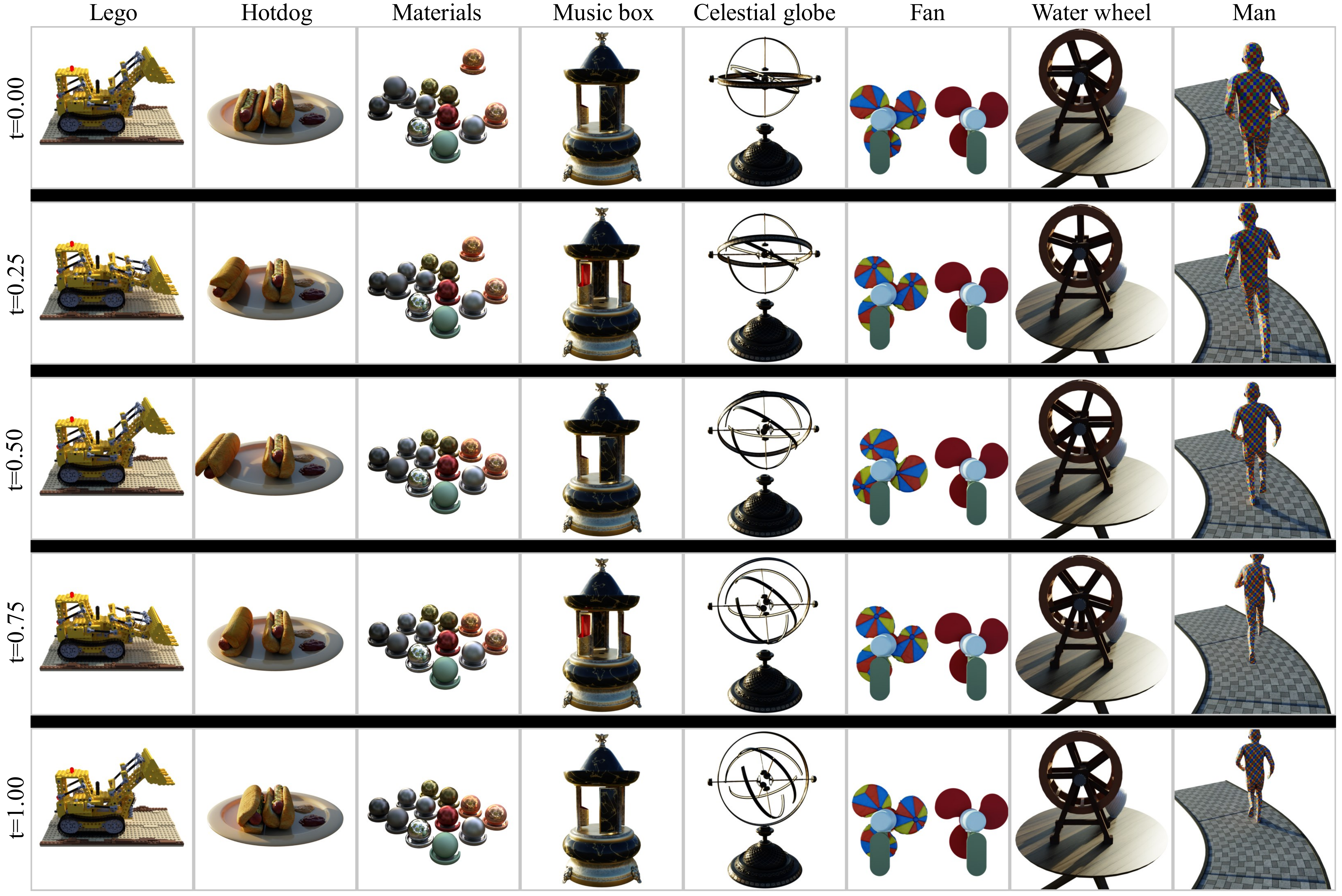}
  \caption{Visualization of our dataset, captured at different time steps under a fixed viewpoint. \textbf{Please see the supplementary video for details.}}
  \label{fig:supp_showing}
\end{figure*}

\noindent\textbf{Scene Motion Visualization.}
\cref{fig:supp_showing} presents all synthetic scenes in our dataset at different time steps under a fixed viewpoint. To better visualize the dynamic content, we recommend watching the supplementary video. Specifically, we developed new animations based on the static scenes provided by NeRF \cite{mildenhall2021nerf} to create the dynamic scenes ``Lego'', ``Hotdog'', and ``Materials''. The ``Music box'' and ``Celestial globe'' are brand-new single-object dynamic scenes. The ``Fan'', ``Water wheel'', and ``Man'' are combination scenes, consisting of independent dynamic and static objects. Notably, the scene ``Fan'' (a rotating electric fan) represents a typical \textit{high-speed} 4D scene, while ``Man'' is another particularly challenging scene due to the \textit{large object displacements} and complex texture variations it involves.

\noindent\textbf{Simulating Threshold Variations.} 
By default, the simulated thresholds $C$ vary between 0.1 and 0.3, encompassing the approximate nominal thresholds of commonly used event cameras, such as the Prophesee Gen 4 \cite{Prophesee_Gen} and DAVIS 346 \cite{DAVIS346}. To simulate threshold variations over polarity, the positive thresholds $C_{p}$ and negative thresholds $C_{n}$ are constrained by the relation: $C_{n} = C_{p} \times n$, where $ n \in \mathcal{N}(\mu,\sigma^{2}) $ with $\mu = 1.0$ and $\sigma = 0.1$. To simulate threshold variations over time, the threshold parameters in the event camera simulator ESIM \cite{rebecq2018esim} are updated every 50 frames. Threshold variations over space are simulated by introducing noisy events. Specifically, Gaussian noise with $\mu = 0$ and $\sigma = 1.0$ is generated, converted to integer values to form discrete noisy events. These noisy events are introduced at a specified fraction, controlled by a masking process to regulate the noise amount.

Please note that \textit{the distribution of simulated threshold variations is \underline{unknown} to our method}. The threshold modeling in Eq.8 of the main paper, relies solely on the event generation model \cite{klenk2023nerf, cannici2024mitigating}, meaning it holds true under any threshold distributions. In other words, Eq.8 is fundamentally data-driven and capable of modeling arbitrary threshold variations. This capability allows our method to remain effective even when encountering the complex threshold distributions in real-world. This is also the reason why we do not follow \cite{delbruck2020v2e} to oversimplify the modeling of threshold variations using a Gaussian distribution.

\subsection{Hybrid Camera System}

Following prior work \cite{cannici2024mitigating, qi2024deblurring, qi2023e2nerf, qi20243, yu2024evagaussians}, we adopt a coaxial camera system instead of a stereo setup. In a stereo system, the optical centers of the event and frame cameras are misaligned, which makes precise alignment challenging even with rigorous calibration. Such misalignment could degrade the performance of both the event rendering loss in Eq.(4) of the main paper and the EDI in \cref{sec:dynamic_blurry} of this supplementary material. Additionally, the real-world scenes captured in prior work \cite{cannici2024mitigating, qi2024deblurring, qi2023e2nerf, qi20243, yu2024evagaussians} suffer from low-resolution and low-quality events. Therefore, we build a hybrid camera system rather than using DAVIS346, in order to capture higher-quality datasets.

As mentioned in the main paper, our hybrid camera system consists of a frame camera (Basler acA1440-220uc with a resolution of 1440 $\times$ 1080 and an 8mm lens), an event camera (Prophesee Gen4 with a resolution of 1280 $\times$ 720 and an 8mm lens), a beam splitter (Thorlabs CCM1-BS013) and a microcontroller (STM32F103C8). Following \cite{wang2023unsupervised}, we performed geometric calibration and temporal synchronization between the event camera and the frame camera.

\noindent\textbf{Geometric Calibration.} 
We first display a blinking chessboard pattern (6 $\times$ 9) on a 27-inch monitor positioned 1 meter in front of the hybrid camera system. The frame camera captures frames at 30 FPS, while the event camera accumulates a one-second event stream to generate an event image. After obtaining both the frames and event images, we apply OpenCV’s corner detection algorithm to extract the chessboard corners. The detected corners are visualized in \cref{fig:supp_calibration}. To ensure accuracy, any incorrectly detected corners are manually removed.

\begin{figure*}[h!]
  \centering
  \includegraphics[width=0.8\linewidth]{figs/Supp/Supp_标定图.pdf}
  \caption{The detected corners for homograpy computation from the frames and event images.}
  \label{fig:supp_calibration}
\end{figure*}

Our goal is to geometrically map pixels from the frame camera view to the corresponding pixels in the event camera view within the overlapping field of view. To achieve this, we employ a homography matrix, estimated using the detected corner correspondences between the two camera views. Formally, this transformation is expressed as,

\begin{equation}
p_{i}^{E}=H\cdot  p_{i}^{F},
\end{equation}

where $H$ denotes a 3 $\times$ 3 transformation matrix, $p_{i}^{E}=[x_{i}^{E},y_{i}^{E},1]^{T}$ and $p_{i}^{F}=[x_{i}^{F},y_{i}^{F},1]^{T}$ are the homogeneous coordinates in the event images and frames. Finally, we warp the frame using the computed matrix H to spatially align the two camera views.

\noindent\textbf{Temporal Synchronization.} 
In our hybrid camera system, two cameras are temporally synchronized by external triggers. A microcontroller acts as the master device, generating a square wave synchronization signal, while the frame and event cameras act as slave devices. On each falling edge of the square wave, the frame camera captures a frame, while the event camera records an external event (logging the synchronization timestamp). The FPS of the frame camera is determined by the frequency of the square wave generated by the microcontroller.

\section{Additional Qualitative Results}
\label{sec:supp_qual_results}

\noindent\textbf{Additional Qualitative Comparisons.} 
In \cref{fig:real_supp_comparison} and \cref{fig:sync_supp_comparison}, we present qualitative comparisons of baselines and our method on the remaining real-world and synthetic scenes, respectively. These visual comparisons highlight the effectiveness of our method in reconstructing high-fidelity dynamic scenes. Specifically, in the ``Lego'' scene of \cref{fig:sync_supp_comparison}, only our approach successfully captures the small raised and recessed dots on the toy's surface, while other methods exhibit noticeable structural distortions and blurring.

\begin{figure*}[h!]
  \centering
  \includegraphics[width=1.0\linewidth]{figs/Supp/Real_定性_Supp.pdf}
  \caption{Qualitative comparisons on the remaining two real-world scenes.}
  \label{fig:real_supp_comparison}
\end{figure*}

\begin{figure*}[h!]
  \centering
  \includegraphics[width=1.0\linewidth]{figs/Supp/Sync_定性_supp.pdf}
  \caption{Qualitative comparisons on the remaining five synthetic scenes.}
  \label{fig:sync_supp_comparison}
\end{figure*}

\clearpage
\newpage

\begin{figure*}[h!]
  \centering
  \includegraphics[width=1.0\linewidth]{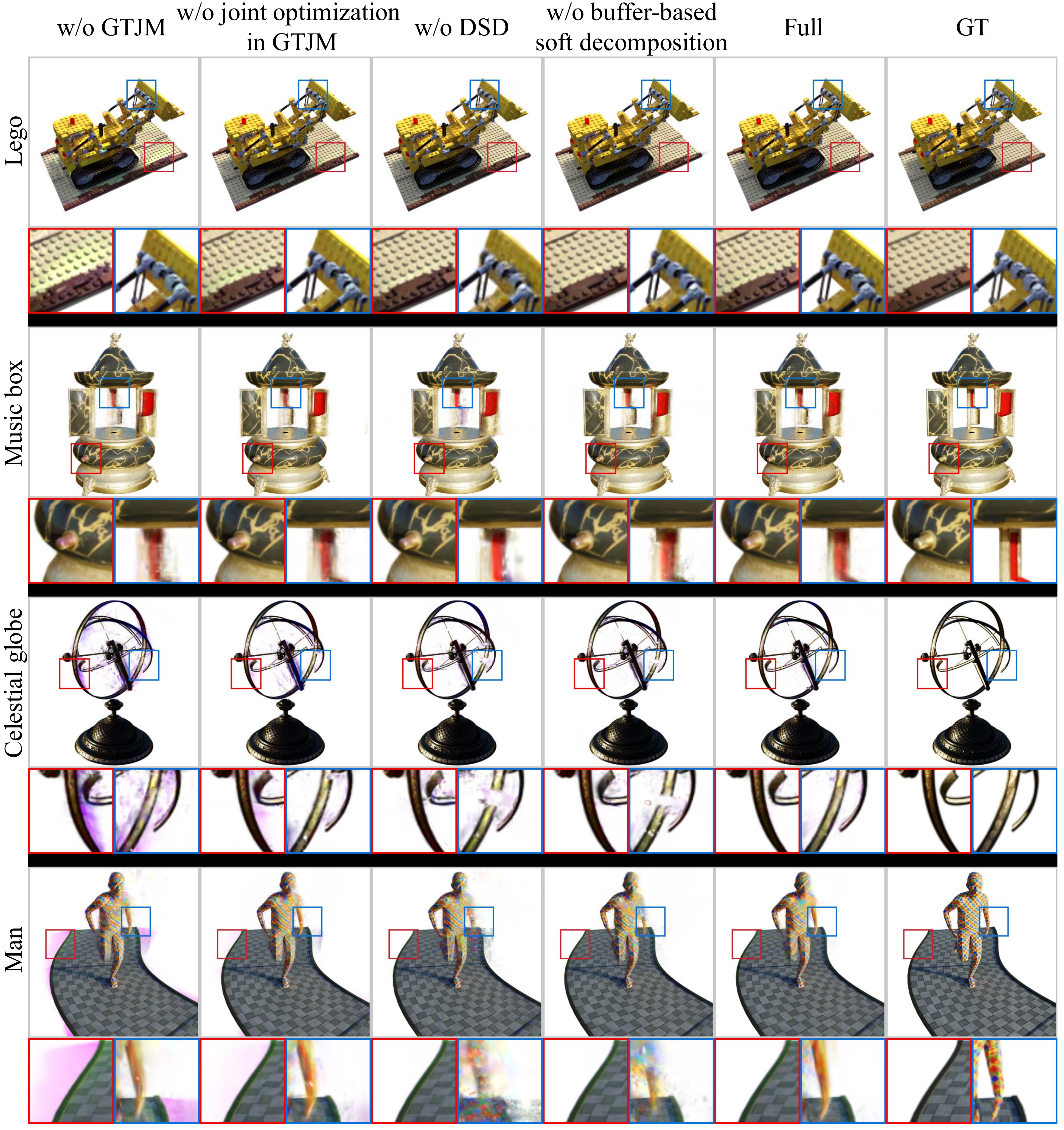}
  \caption{Qualitative results of ablation study on synthetic dataset.}
  \label{fig:supp_ablation}
\end{figure*}

\noindent\textbf{Qualitative Results of Ablation Study.} Additionally, as shown in \cref{fig:supp_ablation}, we provide qualitative results of ablation study on synthetic dataset. Notably, in the subplots with \textcolor[HTML]{FF0000}{red} borders, the absence of GTJM or joint optimization results in color artifacts (e.g., the yellow-green haze in the ``Lego'' scene and the purple haze in the ``Celestial Globe'' scene). In the subplots with \textcolor[HTML]{0072E3}{blue} borders, the absence of DSD and buffer-based soft decomposition leads to deficient and distorted dynamic structures.

\section{More Details on Implementation}
\label{sec:supp_implementation}

\noindent \textbf{The Algorithm.} Our method builds upon the differentiable rasterizer of 3D-GS \cite{kerbl20233d} and is implemented in PyTorch, running on an NVIDIA RTX 3090 GPU. Given sparse RGB frames, we initialize from random point clouds rather than structure-from-motion (SfM) \cite{schonberger2016structure} points. Hyperparameters are provided in the supplementary materials. 

We summarize the detailed training procedure in \cref{alg:training_procedure}. Our method takes sparse RGB frames $I$ and the corresponding event streams $\mathcal{E}$ as input. 

\textbf{Stage 1: RGB-assisted Initial Estimation.} We first leverage the intensity information from the RGB frames to estimate the initial threshold $C$. We initialize $C$ as a constant and optimize it using $\mathcal{L}_{\text{thres}}$ for 7k iterations. 

\textbf{Stage 2: Dynamic-static Decomposition.} Next, we decompose the scene into dynamic and static components. Warm-up Gaussians $GS_{\text{init}}$ are initialized using randomly generated points and optimized for 3k iterations using $\mathcal{L}_{\text{rgb}}$. Through the dynamic-static decomposition pipeline in Section 3.3, we decompose $GS_{\text{init}}$ into static Gaussians $GS_{\text{s}}$ and dynamic Gaussians $GS_{\text{d}}$. 

\textbf{Stage 3: Deformation Warm-up.} We then warm up the deformation field using only RGB frames to accelerate overall training speed. We randomly initialize the deformation field parameters $\theta$ and jointly optimize $GS_{\text{s}}$, $GS_{\text{d}}$, and $\theta$ using $\mathcal{L}_{\text{rgb}}$. 

\textbf{Stage 4: Joint Threshold and GS Optimization.} Finally, we introduce event supervision and perform joint threshold and GS optimization. The total loss $\mathcal{L}_{\text{total}}$ is used to jointly optimize $GS_{\text{s}}$, $GS_{\text{d}}$, $\theta$, and $C$.

\medskip

\noindent\textbf{Hyperparameter Settings.} The learning rate for 3D-GS \cite{kerbl20233d} is initialized at $1.6 \times 10^{-4}$ and decays to $1.6 \times 10^{-5}$ by the end of training. The deformation field, modeled as an MLP network \cite{yang2024deformable}, starts with the same learning rate of $1.6 \times 10^{-4}$, which similarly decreases to $1.6 \times 10^{-5}$. The learnable threshold uses a learning rate of $2.4 \times 10^{-4}$ during the RGB-assisted initial estimation phase, and $1.6 \times 10^{-4}$ during the joint optimization phase, decaying to $1.6 \times 10^{-5}$. The time bin for the ECM is set to 64, enabling the capture of fine-grained threshold variations over time. The weighting factors of the loss functions, $\lambda_{\text{s}}$, $\lambda_{\text{event}}$, and $\lambda_{\text{thres}}$, are set to 0.2, 5.0, and 5.0, respectively.









\begin{algorithm}[h]
\caption{Training for Event-boosted Deformable 3D Gaussians}
\label{alg:training_procedure}
\begin{algorithmic}[1]

\REQUIRE Sparse RGB frames $I$, event streams $\mathcal{E}$

\vspace{0.5em}
\textbf{Stage 1: RGB-assisted Initial Estimation}
\STATE \quad Initialize threshold $C$ to a constant.
\STATE \quad Optimize $C$ for 7k iterations as
\[
\vspace{-0.8em}
C^{\ast } =\underset{C}{\mathrm{argmin}} \mathcal{L}_{\text{thres}}(C, I, \mathcal{E}).
\]

\vspace{0.5em}
\textbf{Stage 2: Dynamic-static Decomposition}
\STATE \quad Randomly initialize static Gaussians $GS_{init}$.
\STATE \quad Optimize $GS_{init}$ for 3k iterations as
\vspace{-0.8em}
\[
\vspace{-0.8em}
GS_{init}^{\ast } =\underset{GS_{init}}{\mathrm{argmin}} \mathcal{L}_{rgb}(GS_{init}, I).
\]

\STATE \quad Decompose $GS_{init}$ into static Gaussians $GS_{s}$ and dynamic Gaussians $GS_{d}$.

\vspace{0.5em}
\textbf{Stage 3: Deformation Warm-up}
\STATE \quad Randomly initialize deformation field parameters $\theta$.
\STATE \quad Joint optimize $GS_{s}$, $GS_{d}$, and $\theta$ for 5k iterations as
\vspace{-0.8em}
\[
\vspace{-0.8em}
GS_{s}^{\ast } , GS_{d}^{\ast }, \theta^{\ast } =\underset{GS_{s}, GS_{d}, \theta}{\mathrm{argmin}} \mathcal{L}_{rgb}(GS_{s}, GS_{d}, \theta, I).
\]

\vspace{0.5em}
\textbf{Stage 4: Joint Threshold and GS Optimization}
\STATE \quad Joint optimize $GS_{s}$, $GS_{d}$, $\theta$, and $C$ for 42k iterations as
\vspace{-0.8em}
\[
\vspace{-0.8em}
GS_{s}^{\ast } , GS_{d}^{\ast }, \theta^{\ast }, C^{\ast } =\underset{GS_{s}, GS_{d}, \theta, C}{\mathrm{argmin}} \mathcal{L}_{total}(GS_{s}, GS_{d}, \theta, C, I, \mathcal{E}).
\]

\end{algorithmic}
\end{algorithm}

\newpage

{
    \small
    \bibliographystyle{ieeenat_fullname}
    \bibliography{main}
}
